\documentclass[10pt,twocolumn,letterpaper]{article}

\usepackage{cvpr}              
\usepackage{multirow}
\usepackage{colortbl}
\usepackage{xcolor}
\definecolor{lightgray}{RGB}{220,220,220}
\usepackage{booktabs}
\usepackage{siunitx}
\usepackage{multirow}
\usepackage[table]{xcolor}

\usepackage{amsmath} 
\usepackage{arydshln}
\usepackage{makecell} 
\usepackage{seqsplit}
\usepackage{graphicx} 
\usepackage[table]{xcolor}
\definecolor{cvprblue}{rgb}{0.21,0.49,0.74}
\definecolor{codegray}{RGB}{240,240,240}
\definecolor{codegreen}{RGB}{0,128,0}
\definecolor{codepurple}{RGB}{153,0,153}

\newcommand{\fn}[1]{\footnotesize{#1}}

\newcommand{\bbf}[1]{\textcolor{cvprblue}{\bf{\fn{#1}}}}

\newcommand{\add}[1]{\textcolor{blue}{\uline{#1}}}
\newcommand{\del}[1]{\textcolor{red}{\sout{#1}}}
\newcommand{\change}[2]{\del{#1}\add{#2}}
\usepackage[accsupp]{axessibility}  

\usepackage[pagebackref,breaklinks,colorlinks,allcolors=cvprblue,urlcolor=black]{hyperref}
\AtBeginDocument{
  \setlength\abovedisplayskip{2pt plus 2pt minus 2pt}
  \setlength\belowdisplayskip{2pt plus 2pt minus 2pt}
  \setlength\abovedisplayshortskip{0pt plus 3pt}
  \setlength\belowdisplayshortskip{3pt plus 2pt minus 2pt}
}
\makeatletter
\newcommand{\makeMySupplementaryTitle}{
    \clearpage
    \onecolumn
    
    \begin{center}
      \Large
      \textbf{\thetitle} \\
      \vspace{0.5em}
      {Supplementary Material} \\
    \end{center}
    
    \vspace{1.0em}
    \setcounter{page}{1}
}
\makeatother

\usepackage{xcolor}
\usepackage{enumitem}
\usepackage{listings}
\usepackage{caption} 
\usepackage[normalem]{ulem}
\usepackage{marvosym}

\definecolor{takeawayborder}{HTML}{E8A978}
\definecolor{takeawaytitlebg}{HTML}{FCDCB6}
\definecolor{takeawaybodybg}{HTML}{FFF5E7}

\usepackage[breakable]{tcolorbox}

\newcolumntype{L}[1]{>{\raggedright\arraybackslash}p{#1}}
\newcolumntype{C}[1]{>{\centering\arraybackslash}p{#1}}
\newcolumntype{R}[1]{>{\raggedleft\arraybackslash}p{#1}}

\def\paperID{5319} 
\def\confName{CVPR}
\def\confYear{2026}

\title{CodePercept: Code-Grounded Visual STEM Perception for MLLMs}

\author{Tongkun Guan\textsuperscript{\rm 1}, Zhibo Yang\textsuperscript{\rm 2\textsuperscript{$\dagger$}}, Jianqiang Wan\textsuperscript{\rm 2}, Mingkun Yang\textsuperscript{\rm 2}, 
Zhengtao Guo\textsuperscript{\rm 3}, Zijian Hu\textsuperscript{\rm 1}, \\
Ruilin Luo\textsuperscript{\rm 4}, Ruizhe Chen\textsuperscript{\rm 5}, Songtao Jiang\textsuperscript{\rm 5}, Peng Wang\textsuperscript{\rm 2\textsuperscript{$\dagger$}}, Wei Shen\textsuperscript{\rm 1(\Letter)}, Junyang Lin\textsuperscript{\rm 2}, Xiaokang Yang\textsuperscript{\rm 1(\Letter)}
\\
\textsuperscript{\rm 1} MoE Key Lab of Artificial Intelligence, AI Institute, School of Computer Science, \\ Shanghai Jiao Tong University
\textsuperscript{\rm 2} Qwen Team
\textsuperscript{\rm 3} Beijing Institute of Technology \\
\textsuperscript{\rm 4} Tsinghua University \textsuperscript{\rm 5} Zhejiang University\\
$^\dagger$Project leader. \quad $^{\textrm{\Letter}}$Corresponding Author. \\
{\tt\small gtk0615@sjtu.edu.cn}
}
\begin{document}
\maketitle
\begingroup 
\renewcommand\thefootnote{} 
\footnotetext{$^\dagger$Project leader. \quad $^{\textrm{\Letter}}$Corresponding Author.} 
\footnotetext{\textbf{Acknowledgements:} This work was supported by the NSFC under Grant 62322604 and 62576207.}
\endgroup 
\begin{abstract}
When MLLMs fail at visual reasoning, a fundamental question arises: is it due to perceptual deficiencies or reasoning limitations? Through systematic scaling analysis that independently scales perception and reasoning components, we uncover a critical insight: scaling perception consistently outperforms scaling reasoning. This reveals perception as the true lever limiting current STEM visual reasoning. Motivated by this insight, our work focuses on systematically enhancing the perception capabilities of MLLMs by establishing code as a powerful perceptual medium—executable code provides precise semantics that naturally align with the structured nature of STEM visuals. Specifically, we construct ICC-1M, a large-scale dataset comprising 1M Image-Caption-Code triplets that materializes this code-as-perception paradigm through two complementary approaches: (1) Code-Grounded Caption Generation treats executable code as ground truth for image captions, eliminating the hallucinations inherent in existing knowledge distillation methods; (2) STEM Image-to-Code Translation prompts models to generate reconstruction code, mitigating the ambiguity of natural language for perception enhancement. To validate this paradigm, we further introduce STEM2Code-Eval, a novel benchmark that directly evaluates visual perception in STEM domains. Unlike existing work relying on problem-solving accuracy as a proxy that only measures problem-relevant understanding, our benchmark requires comprehensive visual comprehension through executable code generation for image reconstruction, providing deterministic and verifiable assessment.  Code is available at \url{https://github.com/TongkunGuan/Qwen-CodePercept}.
\end{abstract}    
\section{Introduction}
\label{sec:intro}
Recent breakthroughs in reinforcement learning have triggered an ``Aha moment"~\cite{guo2025deepseek,shao2024deepseekmath} for Large Language Models (LLMs), inspiring extensive research efforts to replicate this success in multimodal domains~\cite{team2025kwai,team2025kimi,bai2025intern,team2025minicpm4,xiaomi2025mimo,wang2025internvl3,jiang2025hulu,jiang2024med}. This trend is particularly prominent in Science, Technology, Engineering, and Mathematics (STEM), where researchers employ staged learning and sophisticated reward mechanisms to unlock stronger cross-modal reasoning of MLLMs. However, a fundamental question remains: \textbf{\emph{What is the true bottleneck limiting MLLMs in STEM, and when models fail, is it due to perception deficiencies or reasoning limitations?}}

\begin{figure}[t]
    \centering
    \includegraphics[width=\linewidth]{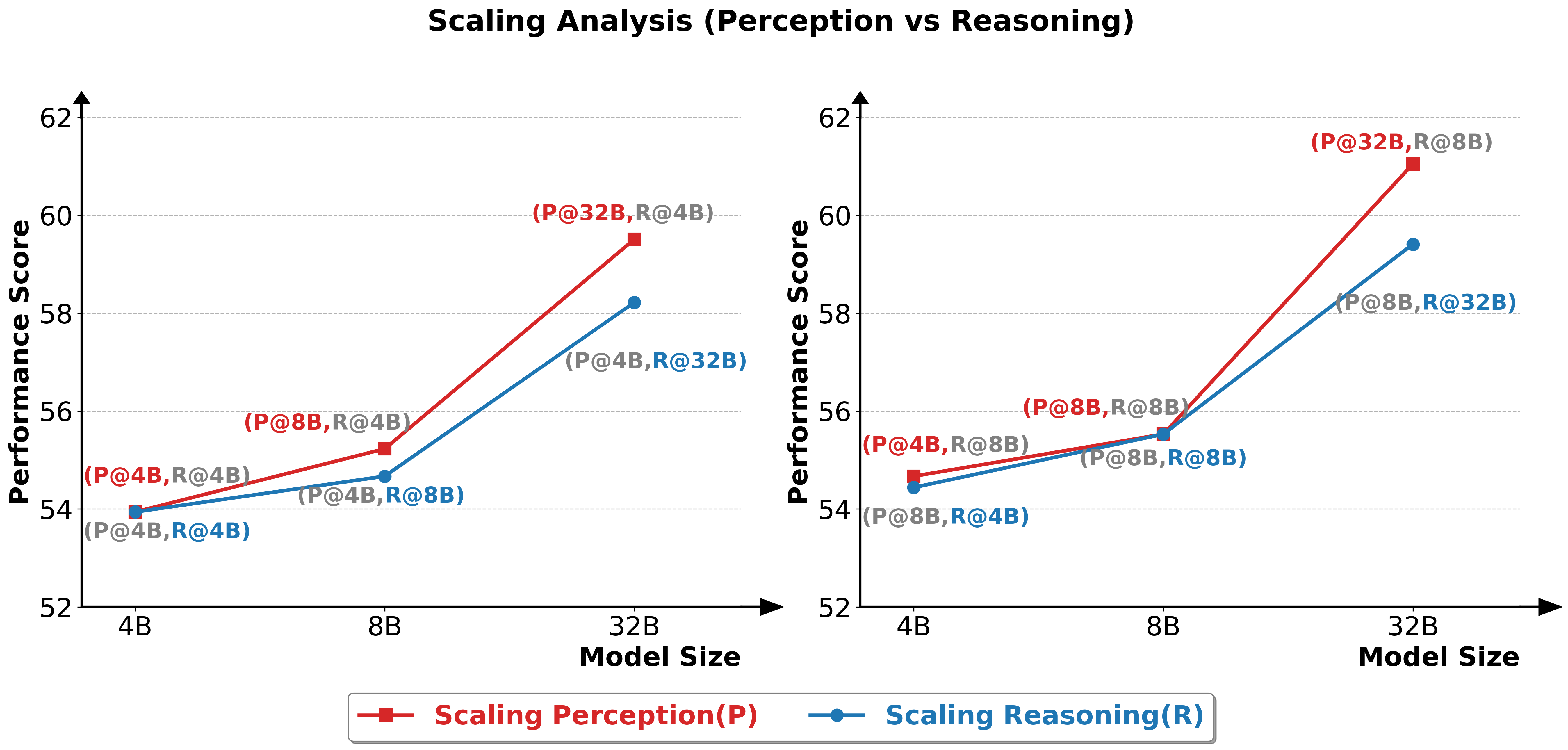}
    \vspace{-1.5em}
    \caption{The scaling analysis reveals perception as the bottleneck in STEM. We decouple visual STEM reasoning into perception (image-to-caption) and reasoning (caption-to-answer) stages, then independently scale each component while keeping the other constant. All components use Qwen3-VL-Thinking~\cite{Qwen3VL_github} models and evaluation in the representative MathVision Dataset~\cite{MathVision}. Both experiments demonstrate that scaling perception consistently outperforms scaling reasoning. This finding motivates our focus on systematically enhancing MLLMs' perception capabilities in STEM.
    }
    \label{fig:P&R}
    \vspace{-1em}
\end{figure}
    
To answer this question, we conduct a systematic scaling analysis by decoupling the task into two stages: visual perception (generating image descriptions) and reasoning (solving problems based solely on textual descriptions). 
We independently scale perception and reasoning capabilities while holding the other fixed, as illustrated in Fig.~\ref{fig:P&R}. 
The results demonstrate that expanding perception capacity consistently yields greater performance gains than expanding reasoning capacity.  This empirical evidence illuminates a critical insight: \textbf{\emph{perception is the true lever that unlocks current STEM visual reasoning.}}

Motivated by this insight, we focus this work on systematically enhancing the perception capabilities of MLLMs for STEM domains. A seemingly intuitive solution would be to enhance STEM perception through knowledge distillation, leveraging advanced MLLMs like GPT-series or Gemini-series to generate descriptive captions for training. However, this path encounters critical limitations. First, these teacher models are prone to hallucination, producing factually incorrect descriptions, particularly regarding spatial positioning, quantitative relationships, and element interactions. Second, and more critically, many STEM images exhibit what we term \emph{descriptive aphasia}: their complex spatial relationships and precise numerical values cannot be fully or accurately captured by natural language alone. For instance, precisely describing auxiliary line constructions in complex polyhedral geometry remains inherently challenging for natural language.

Furthermore, the field lacks a direct paradigm for evaluating visual perception capabilities in STEM domains. Existing research~\cite{lu2025omnicaptioner} predominantly relies on problem-solving accuracy as a proxy for evaluating perceptual ability, yet this metric only reflects the model's capacity to understand problem-relevant information rather than measuring true comprehensive visual comprehension.

These limitations motivate a paradigm shift: \textbf{\emph{grounding perception in executable code.}}
We argue that requiring MLLMs to generate executable Python code for image reconstruction offers the most rigorous validation of perceptual capabilities. The principle is straightforward yet powerful: accurate image reproduction is possible only when a model achieves complete visual comprehension.

Building on this principle, we introduce \textbf{STEM2Code-Eval}: a manually annotated benchmark of 1,000 images that challenges models to generate executable Python code for image reconstruction, providing deterministic and verifiable assessment of visual perception. STEM2Code-Eval draws images from established STEM benchmarks~\cite{DynaMath, LogicVista,MathVerse,MathVision,MathVista,WeMath}, encompassing diverse domains including mathematics, physics, chemistry, and electrical engineering. We employ a rigorous three-stage pipeline combining code agent generation, candidate selection, and human annotation to ensure code quality.

Beyond benchmarking, we further propose code as a powerful medium to significantly enhance MLLMs' visual perception capabilities through two Code-Grounded tasks: \textbf{1) Code-Grounded Caption Generation}, which leverages executable code as ground truth for generating image captions, effectively eliminating AI-generated description errors; and \textbf{2) STEM Image-to-Code Translation}, which directly trains models to generate executable reconstruction code, removing the inherent ambiguity of natural language descriptions. 
To enable effective training for these Code-Grounded tasks, we construct \textbf{ICC-1M}: a large-scale training dataset containing over 1M high-quality STEM \textbf{I}mage-\textbf{C}aption-\textbf{C}ode pairs. Our data construction employs a synthesis strategy with three pipelines: 1) executable Python code to image reproduction, 2) principled diversification that extracts STEM concepts from seed images and re-instantiates them across diverse visual contexts while preserving conceptual validity, and 3) specialized solid geometry synthesis to address the fundamental limitations of current MLLMs in generating solid geometry codes. Through rigorous three-stage quality control, ICC-1M provides a robust foundation for training models to enhance visual perception and executable code generation in STEM domains. In summary, our contributions are threefold: 
\begin{itemize}
    \item We identify perception is the primary bottleneck in STEM visual reasoning through rigorous scaling analysis; 
    \item We introduce STEM2Code-Eval, a manually curated benchmark that establishes code generation as a verifiable ground truth for evaluating visual perception in STEM;
    \item We construct ICC-1M to propose two Code-Grounded training tasks that systematically enhance perceptual capabilities. Experiments demonstrate the effectiveness and prove that code is an alternative to caption.
\end{itemize}

\section{Related Work}
\noindent \textbf{MLLMs for STEM}
Current MLLMs have predominantly focused on enhancing reasoning capabilities to address STEM-related challenges~\cite{team2025kwai,team2025kimi,bai2025intern,team2025minicpm4,xiaomi2025mimo,wang2025internvl3,lu2025ovis2,hong2025glm,bai2025qwen2,guo2025seed1,guan2025token,wang2025marten,jiang2025capo,chen2025datasets,wang2024qwen2,bai2023qwen,wang2025mathcoder,shi2025mathcanvas}. Recent advances in this direction can be categorized into three main methods: 1) Cold-start thinking data curation, where researchers meticulously design high-quality seed datasets that provide reasoning patterns~\cite{huang2025vision,meng2025mm,deng2025openvlthinker,wang2025vl}; 2) RL-based methods, which employ reinforcement learning with carefully designed reward mechanisms~\cite{zhang2025r1,yang2025r1,yu2025perception,luo2025ursa} to iteratively improve reasoning performance; 3) Unimodal thinking data transfer, where several studies~\cite{wei2025open,chen2025advancing,xiaomi2025mimo} demonstrate that high-quality text-only reasoning data can substantially enhance the reasoning capabilities of MLLMs when appropriately integrated. 
Despite these significant advances in reasoning, our scaling analysis reveals that visual perception remains the primary bottleneck in STEM field. However, existing research has largely overlooked the fundamental role, with few works explicitly addressing perceptual deficiencies. To bridge this gap, we introduce code as a verifiable ground truth for comprehensive visual understanding and developing systematic training methodologies to enhance perceptual capabilities through executable code.

\noindent \textbf{STEM Visual Perception Evaluation}
Evaluating visual perception capabilities in STEM domains presents unique challenges for MLLMs. This includes accurately perceiving element types, quantities, structures, relationships, and underlying principles~\cite{guan2024bridging,guan2023self_,guan2023self,guan2025posformer,guan2022industrial,fu2025multimodal,guan2025ccdplus}. A recent work~\cite{lu2025omnicaptioner} adopt a two-stage evaluation paradigm where image captioning is followed by LLM-based problem solving to isolate perceptual capabilities. However, this approach only measures problem-relevant information extraction rather than comprehensive visual understanding, potentially overlooking critical visual details that are irrelevant to specific questions but essential for complete perception. More importantly, different from domain-specific (UI~\cite{si2025design2code,yun2024web2code,lin2023designbench,ge2025advancing}, Chart~\cite{yang2024chartmimic}, SVG~\cite{qiu2024can,li2025unisvg,yang2025omnisvg}) code generation tasks that primarily target downstream applications, we develop STEM Image-Code pairs serves dual purposes: (1) establish a rigorous benchmark for evaluating comprehensive visual perception, and (2) enable the construction of high-quality image-code-caption triplets for perception enhancement.

\section{CodePercept Methodology}
\label{sec:methodology}
\begin{figure*}
    \centering
    \vspace{-1em}
    \includegraphics[width=0.9\linewidth]{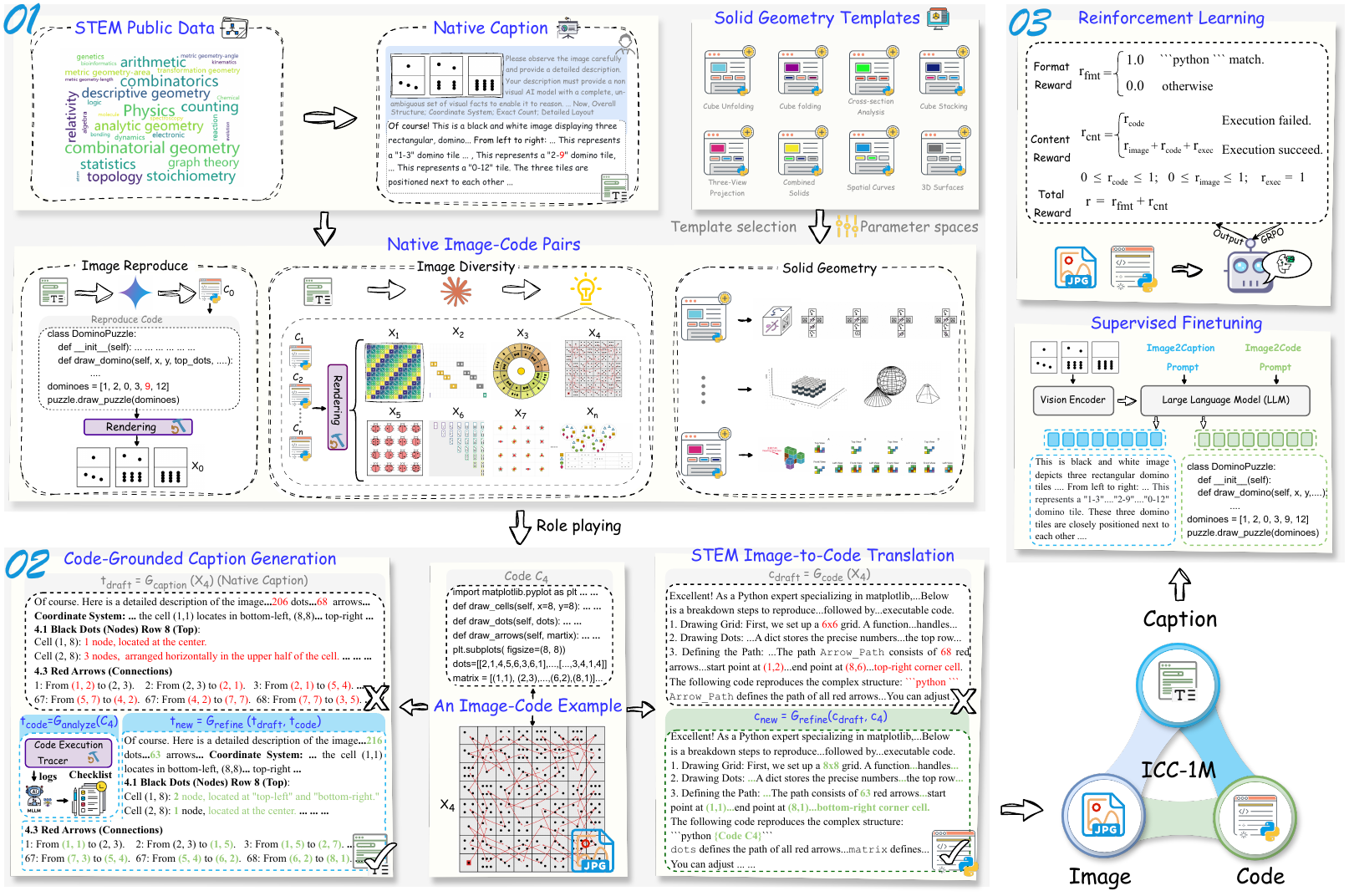}
    \vspace{-1em}
    \caption{The overview pipeline of CodePercept that enhances MLLMs' visual perception in STEM domains through code-grounded learning. \textbf{(Part 01)} Starting from public STEM data, we construct high-quality image-code pairs via three complementary pipelines: (1) \textit{Image Reproduce} converts existing STEM images into executable Python codes, (2) \textit{Image Diversity} extracts concepts from seed images and generates diverse instantiations while preserving semantic validity, and (3) \textit{Solid Geometry} employs parametric templates to generate complex solid geometry images with corresponding codes, addressing MLLMs' solid geometry limitations. \textbf{(Part 02)} The synthesized data enables two novel training tasks (Code-Grounded Caption Generation and STEM Image-to-Code Translation) that fundamentally shift how we approach visual perception. \textbf{(Part 03)} These processes culminate in ICC-1M, a dataset of over 1M curated image-caption-code triplets. We employ both Supervised Finetuning and Reinforcement Learning to train models that achieve robust visual perception capabilities.}
    \vspace{-1em}
    \label{fig:codepercept}
\end{figure*}
As illustrated in Fig.\ref{fig:codepercept}, we construct native STEM image-code pairs using the matplotlib library~\cite{tosi2009matplotlib} to support subsequent code-grounded caption generation and STEM image-to-code translation tasks. We then enhance the model's perceptual capabilities by the supervised fine-tuning stage and reinforcement learning stage. For clarity, the following sections describe the complete CodePercept pipeline.

\subsection{Image-Code Pair Construction}
\label{subsec:data_engine}
Given a image $\mathbf{I}$, we leverage MLLMs to generate corresponding reproduction code $\mathbf{c}$ and render it to produce $\mathbf{x}$. While MLLMs may not achieve perfect reconstruction (\emph{i.e.}, $\mathbf{x} \neq \mathbf{I}$), we observe that the generated image-code pairs $(\mathbf{x}, \mathbf{c})$, after rigorous validation, exhibit high consistency between visual output and code semantics, as well as strong STEM correctness in representing the core STEM concepts.

Building on this insight, we design a scalable data engine that expands a seed dataset $\mathcal{X}$ of STEM-focused public training images into a large-scale image-code dataset through three parallel pipelines (Fig.\ref{fig:codepercept}, Part 01):
\begin{equation} 
\{(\mathbf{x},\mathbf{c}) | \mathbf{c} \in \texttt{F}_{\text{IR}}(\mathcal{X}) \cup \texttt{F}_{\text{ID}}(\mathcal{X}) \cup \texttt{F}_{\text{SG}}(\mathcal{T}_{\text{geo}}), \mathbf{x} \sim \mathcal{R}(\mathbf{c})\}
\end{equation} 
where $\texttt{F}_{\text{IR}}$, $\texttt{F}_{\text{ID}}$, and $\texttt{F}_{\text{SG}}$ represent image reproduction, image diversity, and solid geometry synthesis pipelines, $\mathcal{T}_{\text{geo}}$ refers to a collection of solid geometric templates. $\mathcal{R}(\mathbf{c})$ denotes executing the code to render image $\mathbf{x}$.

\noindent \emph{1) Image Reproduction $\texttt{F}_{\text{IR}}(\cdot)$:} For each $\mathbf{I} \in \mathcal{X}$, we decompose the task into two stages to generate the code: 
\begin{equation}
\mathbf{c} = \texttt{G}_{\text{code}}(\mathbf{I}, \texttt{G}_{\text{caption}}(\mathbf{I}))
\end{equation}
where $\texttt{G}_{\text{caption}}$ prompts a MLLM to extract a rich textual description from the STEM image. The code generation $\texttt{G}_{\text{code}}$ then prompts the MLLM to produce executable Python code $\mathbf{c}$ conditioned on both the image and generated caption. This allows the prompted MLLM to first understand visual content explicitly before generate code, improving both accuracy and interpretability. However, this strategy remains inherently constrained by the diversity of source images in existing datasets, motivating our diversification approach.

\noindent \emph{2) Image Diversity $\texttt{F}_{\text{ID}}(\cdot)$} operates on a fundamental insight: principles underlying STEM images can be systematically abstracted and re-instantiated across different contexts while preserving conceptual validity. This allows us to distill advanced MLLMs' language-to-code capabilities to render more novel images beyond existing datasets. Formally, for each $\mathbf{I} \in \mathcal{X}$, we generate the diversity codes by a compositional abstraction-instantiation process:
\begin{equation}
[\mathbf{c}_1, \mathbf{c}_2, \ldots, \mathbf{c}_K] = \texttt{G}_{\text{code}}(\mathbf{I}, \texttt{G}_{\text{principle}}(\mathbf{I}))
\end{equation}
where $\texttt{G}_{\text{principle}}$ prompts a MLLM to extract the underlying scientific principle from $\mathbf{I}$. $\texttt{G}_{\text{code}}$ prompts the MLLM to generate $K$ diverse code variations based on that principle.
For instance, from a seed image depicting domino-based logic puzzles, our pipeline generates circular domino wheel pattern, triangular domino combination arrangement, ladybug spot matrix, grid connection graph, \emph{etc.}, each maintaining STEM rigor while introducing structural novelty.

\noindent \emph{3) Solid Geometry Synthesis $\texttt{F}_{\text{SG}}(\mathcal{T}_{\text{geo}})$.} Both the reproduction and diversity pipelines face a fundamental limitation when generating code for solid geometry, as this task requires a precise configuration of spatial relationships—a capability where current LLMs and MLLMs are notably deficient.
To address this gap, we construct a code template-based synthesis pipeline that generates solid geometry images.
Formally, we define a collection of code templates\footnote{These templates cover typical solid geometry scenarios, including cube net unfolding and folding sequences, orthographic three-view projection and reconstruction, cross-sectional analysis of solids, cube stacking configurations, combinations of various geometries, polyhedral constructions, spatial curve visualization, and surface integral representations.} $\mathcal{T}_{\text{geo}} = \{\tilde{\mathbf{c}}_i\}_{i=1}^{M}$, where each template $\tilde{\mathbf{c}}_i$ defines the generation logic for a specific geometric type and is parameterized by a parameter space $\Theta_i$ to control attributes such as cube arrangements, viewing angles, and spatial configurations. The synthesis process is formulated:
\begin{equation}
 \mathcal{C}_{\text{geo}} = \{\mathbf{c}_{i} \mid \mathbf{c}_{i} = \tilde{\mathbf{c}}_i(\boldsymbol{\theta}); i \in {1, \ldots, M}; \boldsymbol{\theta} \in \Theta_i\}
\end{equation}
where we instantiates each template $\tilde{\mathbf{c}}_i$ with parameters $\boldsymbol{\theta}$ sampled from $\Theta_i$ to produce code  $\mathbf{c}_i$. 
This ensures geometric correctness through structured templates while achieving visual diversity through systematic parameter sampling.

\noindent \emph{4) Unified Quality Control.}
Despite our carefully designed engines, we cannot guarantee optimal quality for all generated pairs. We therefore develop a composite filtering strategy that retains only those pairs satisfying quality criteria: 
\begin{equation} 
\mathcal{D} = \{(\mathbf{x}, \mathbf{c}) \mid (\mathbf{x}, \mathbf{c}) \in Q_{\text{I}}(\mathcal{X}) \wedge Q_{\text{C}}(\mathcal{C}) \wedge Q_{\text{IC}}(\mathcal{X}, \mathcal{C})\}
\end{equation} 
where $Q_{\text{I}}$ for image quality, $Q_{\text{C}}$ for code quality, and $Q_{\text{IC}}$ for image-code consistency. $\mathcal{X}$ and $\mathcal{C}$ represent the collection of images and code gathered from the previous three pipelines, respectively.
All metrics are evaluated using a SOTA MLLM with specialized prompts. 

\subsection{Code-Grounded Caption Generation}
\label{subsec:code_grounded_caption}
While the quality-controlled image-code pairs provide precise structural representations, accurate natural language descriptions remain essential for comprehensive MLLM training (both validated as complementary modalities in our experiments). Specifically, natural language captions provide semantic understanding while code serves as structured captions with precise spatial and quantitative details.

However, generating semantically accurate captions for complex STEM images presents a fundamentally challenge. Intricate STEM relationships, precise spatial configurations, and abundant quantitative details are all prone to hallucination in direct vision-to-text generation. To solve this problem, we introduce a novel framework that leverages ground-truth code $\mathbf{c}$ as a reliable medium for producing both accurate and linguistically natural descriptions $\mathbf{t}_{\text{new}}$:
\begin{equation}
\mathbf{t}_{\text{new}} = \texttt{G}_{\text{refine}}\left(\texttt{G}_{\text{caption}}(\mathbf{x}), \texttt{G}_{\text{analyze}}(\mathbf{c}, \xi(\mathbf{c}))\right)
\end{equation} 
where $\texttt{G}_{\text{caption}}$ first generates a linguistically natural but potentially inaccurate draft $\mathbf{t}_{\text{draft}}$, $\texttt{G}_{\text{analyze}}$ then extracts verified visual facts from code, and $\texttt{G}_{\text{refine}}$ finally synthesizes factually precise captions while preserving linguistic naturalness.
(Fig.~\ref{fig:codepercept}, Part 02, Left side)

\noindent \emph{1) Native Caption.} We begin by obtaining a description draft $\mathbf{t}_{\text{draft}} = \texttt{G}_{\text{caption}}(\mathbf{x})$ by prompting a MLLM to describe the image directly.  $\mathbf{t}_{\text{draft}}$ exhibits natural language flow but captures factual inaccuracies regarding quantitative details, spatial relationships, and STEM properties.

\noindent \emph{2) Code Analysis.}  
To extract reliable visual facts, we leverage the generation code $\mathbf{c}$ itself as ground truth—since it directly specifies rendered content, it naturally contains definitive information about all visual primitives. However, directly analyzing complex code with deep recursion, nested loops, or intricate transformations remains prohibitively difficult for LLMs.
To address this, we augment code analysis with execution-based verification:
\begin{equation}
\mathbf{t}_{\text{code}} = \texttt{G}_{\text{analyze}}(\mathbf{c}, \xi(\mathbf{c})) = \text{LLM}(\mathbf{c}, \xi(\mathbf{c}), \texttt{P}_{\text{analyze}})
\end{equation}
where $\texttt{P}_{\text{analyze}}$ denotes specialized prompts that guide the LLM to extract visual facts from both the code structure itself and its execution logs $\xi(\mathbf{c})$. The execution tracer $\xi(\mathbf{c})$ functions as an external instruction manual of the code, which captures a structured checklist of logs about all visual elements rendered during code execution. 

Specifically, the execution tracer operates by executing the code in a controlled environment and systematically records geometric precision (exact coordinates, dimensions, spatial relationships), quantitative attributes (definitive counts, RGB specifications), rendering semantics (z-order layering, transformation matrices), and STEM mappings between parameters and visual manifestations. When confronted with complex algorithmic logic, this tracer serves as a definitive reference that resolves ambiguities about spatial orientations, exact quantities, and precise visual attributes. By combining the code structure with the tracer's output, the LLM gains complete and verified information about the image content, thereby eliminating uncertainties inherent in analyzing code in isolation.

\noindent \emph{3) Code-Grounded Caption.} A factuality-preserving rewriting that keeps the complementary strengths in 1) and 2):
\begin{equation}
\mathbf{t}_{\text{new}} = \texttt{G}_{\text{refine}}(\mathbf{t}_{\text{draft}}, \mathbf{t}_{\text{code}}) = \text{LLM}(\mathbf{t}_{\text{draft}}, \mathbf{t}_{\text{code}}, \texttt{P}_{\text{refine}})
\end{equation}
where $\texttt{P}_{\text{refine}}$ instructs an LLM to perform surgical edits that systematically correct factual errors (incorrect numbers, positions, colors, or geometric relationships), replace vague quantifiers with exact counts, and supplement omissions of perceptually salient content. Crucially, these edits maintain the original syntactic structures, language style, and descriptive flow of $\mathbf{t}_{\text{draft}}$ while substituting verified visual facts from $\mathbf{t}_{\text{code}}$, ensuring that the final captions achieve both STEM rigor and pedagogical accessibility. 

\subsection{STEM Image-to-Code Translation}
\label{subsec:image_to_code}
Beyond natural language, code offers a fundamentally different modality for visual description. It defines visual content as programming constructs, capturing geometric relationships, mathematical constraints, and structural details that natural language descriptions often leave ambiguous or incomplete. However, providing ground-truth code alone is insufficient for effective learning. Models need explicit guidance to learn visual-to-code mappings: mapping observed features to code segments and understanding how parameters govern visual properties. To enable this, we construct the explanatory-style code as $\mathbf{c}_{\text{new}} = \texttt{G}_{\text{refine}}(\texttt{G}_{\text{code}}(\mathbf{x}), \mathbf{c})$,
where $\mathbf{c}_{\text{new}}$ represents the final explanatory code with code correctness and instructional richness.
(Fig.~\ref{fig:codepercept}, Part 02, Right side)

\noindent \emph{1) Explanatory Draft Generation.} We first prompt a MLLM to generate code directly from the image: $\mathbf{c}_{\text{draft}} = \texttt{G}_{\text{code}}(\mathbf{x})$. This draft naturally exhibits desirable pedagogical patterns including step-by-step breakdown, explicit parameter choices, and instructional commentary explaining visual-to-code mappings. However, without access to ground-truth specifications, $\mathbf{c}_{\text{draft}}$ frequently contains factual errors in coordinates, dimensions, loop logic, or algorithmic structure, particularly for complex STEM images.

\noindent \emph{2) Code-Grounded Refinement.} To correct these errors while preserving explanatory richness, we leverage the verified ground-truth code $\mathbf{c}$ as a reliable reference:
\begin{equation} \mathbf{c}_{\text{new}} = \texttt{G}_{\text{refine}}(\mathbf{c}_{\text{draft}}, \mathbf{c}) = \text{LLM}(\mathbf{c}_{\text{draft}}, \mathbf{c}, \texttt{P}_{\text{refine}}) \end{equation}
where $\texttt{P}_{\text{refine}}$ instructs an LLM to adaptively refine the explanatory content and replace erroneous code with $\mathbf{c}$, while carefully maintaining the original explanatory structure. This refinement preserves the logical flow, step-by-step explanations, and contextual descriptions of $\mathbf{c}_{\text{draft}}$ while ensuring code correctness through alignment with $\mathbf{c}$.

Explanatory image-code pairs provide rich training signals that teach models not only what code to generate, but why specific implementations best capture content, ensuring both technical precision and instructional effectiveness. 

\begin{table*}[t]
    \centering
    \caption{Evaluating the \textbf{perception abilities} of various MLLMs using a captioner-solver setup. Each MLLM (captioner)  generates an image description, and a fixed LLM solver then performs problem-solving based solely on that description. More accurate descriptions are expected to yield better reasoning results.}
    \vspace{-1em}
    \label{tab:main_results}
    \scalebox{0.9}{
    \begin{tabular}{l L{14mm}L{13mm}L{13mm}L{13mm}L{13mm}L{15mm} L{13mm}}
        \toprule[1.2pt]
        \multirow{2}{*}{\emph{\textbf{Image Captioner}}} & \multicolumn{6}{c}{Benchmark Datasets (\%)} & {\multirow{2}{*}{Average}} \\
        \cmidrule(lr){2-7}
        & {MathVision} & {MathVista} & {MathVerse} & {DynaMath} & {WeMath} & {LogicVista} & \\
        \midrule[1.2pt]
        \multicolumn{8}{c}{\emph{\textbf{LLM Solver: Qwen3-30A3-Thinking~\cite{yang2025qwen3}}}} \\
        \midrule[1.2pt]
        Claude-Opus 4.1-Thinking~\cite{anthropic2024claude} &59.61  &71.10  &56.19  &73.25  &44.86  &59.28 &60.72 \\
        GPT5-Thinking~\cite{openai2023gpt5} &60.03  &65.20  &69.56  &71.00  &54.57  &53.02  & 62.23  \\
        Gemini2.5-Pro &66.80  &74.80  &73.47  &81.42  &60.29  &66.44  &70.53 \\
        \midrule
        KeyeVL1.5-8B~\cite{team2025kwai} &54.11&64.90&49.95&62.37&33.62&45.19&51.69\\
        Intern-S1-8B~\cite{bai2025intern} &51.67&65.70&51.90&63.61&33.43&51.23 &52.92\\
        GLM-4.1V-9B~\cite{hong2025glm} &53.75&64.60&54.47&66.17&40.76&51.00&55.13\\
        InternVL3.5-8B~\cite{wang2025internvl3} &53.32&67.70&53.40&68.12&41.05&51.68&55.88\\
        MiniCPM-V-4.5~\cite{team2025minicpm4} &53.15&66.60&57.84&65.44&43.71&52.57&56.55\\
        Qwen2.5-VL-72B~\cite{bai2025qwen2} &54.14&67.50&55.40&68.28&44.86&52.34&57.09\\
        Qwen3-VL-30A3B-Instruct~\cite{Qwen3VL_github} &53.59&68.00&66.44&71.67&46.10&53.69&59.92\\
        Qwen3-VL-235A22B-Instruct~\cite{Qwen3VL_github} &60.43&73.80&70.08&77.39&53.05&59.73&65.75\\
        \cmidrule(lr){1-8}
        Qwen3-VL-4B-Instruct~\cite{Qwen3VL_github} &54.21 &67.30 &64.59 &69.40 &46.10 &54.14 &59.29 \\
        \rowcolor{cyan!5} \bfseries{CodePercept-4B-S1} &57.63\bbf{+3.4}  &69.60\bbf{+2.3}  &65.59\bbf{+1.0}  &71.38\bbf{+2.0}  &47.81\bbf{+1.7}  &60.40\bbf{+6.3}   & \bfseries{62.07}\bbf{+2.8} \\
        Qwen3-VL-8B-Instruct~\cite{Qwen3VL_github} &54.37  & 69.60 & 63.75 & 72.19 & 45.43 & 56.82 & 60.36 \\
        \rowcolor{cyan!5} \bfseries{CodePercept-8B-S1} &59.31\bbf{+5.0}  &70.20\bbf{+0.6}  &66.52\bbf{+2.8}  &73.20\bbf{+1.0}  &49.14\bbf{+3.7}  &61.52\bbf{+4.7}  &\bfseries{63.32}\bbf{+3.0} \\
        \rowcolor{cyan!5} \bfseries{CodePercept-8B-S1-PAMI} &61.74 &72.60 &69.79 &76.85 &51.90 &66.44 &66.55 \\
        Qwen3-VL-32B-Instruct~\cite{Qwen3VL_github} &58.55&72.20&71.09&75.78&48.00&62.19&64.63\\
        \rowcolor{cyan!5} \bfseries{CodePercept-32B-S1} &62.27\bbf{+3.7}  &72.90\bbf{+0.7}  &71.70\bbf{+0.6}  &77.41\bbf{+1.6}  &54.19\bbf{+6.2}  &65.33\bbf{+3.1}  &\textbf{67.30}\bbf{+2.7} \\
        \bfseries{Caption-32B-S1-PAMI} &64.57\bbf{+3.7}  &73.50\bbf{+0.7}  &73.10\bbf{+0.6}  &78.90\bbf{+1.6}  &56.10\bbf{+6.2}  &66.89\bbf{+3.1}  &\textbf{68.84}\bbf{+2.7} \\
        \midrule[1.2pt]
        \midrule[1.2pt]
        \multicolumn{8}{c}{\emph{\textbf{LLM Solver: Qwen3-235A22-Thinking~\cite{yang2025qwen3}}}} \\
        \midrule
        Qwen3-VL-4B-Instruct~\cite{Qwen3VL_github} &59.80 &69.20 &66.39 &71.22 &48.86 &56.82 &62.05 \\
        \rowcolor{cyan!5} \bfseries{CodePercept-4B-S1} &64.71\bbf{+4.9}  &71.30\bbf{+2.1}  &66.73\bbf{+0.3}  &72.40\bbf{+1.2}  &50.00\bbf{+1.1}  &64.65\bbf{+7.8}  & \bfseries{64.97}\bbf{+2.9} \\
        Qwen3-VL-8B-Instruct~\cite{Qwen3VL_github} &59.67 &71.00 &63.88 &73.69 &49.14 &58.16 &62.59  \\
        \rowcolor{cyan!5} \bfseries{CodePercept-8B-S1} &66.45\bbf{+6.8}  &71.40\bbf{+0.4}  &67.95\bbf{+4.1}  &75.05\bbf{+1.4}  &52.29\bbf{+3.2}  &62.64\bbf{+4.5}  &\textbf{65.96}\bbf{+3.4} \\
        Qwen3-VL-32B-Instruct~\cite{Qwen3VL_github} &62.66 &74.00 &69.90 &75.54 &56.48 &66.44 &67.50\\
        \rowcolor{cyan!5} \bfseries{CodePercept-32B-S1} &69.96\bbf{+7.3}  &75.90\bbf{+1.9}  &73.56\bbf{+3.6}  &79.50\bbf{+4.0}  &57.81\bbf{+1.3}  &70.02\bbf{+3.6}  &\textbf{71.13}\bbf{+3.6} \\
        \bottomrule[1.2pt]
    \end{tabular}
    }
    \label{two-stage math}
\end{table*}

\subsection{Post-Training Strategy}
\label{subsec:training}
Building on these strategies (Sec.~\ref{subsec:code_grounded_caption} and~\ref{subsec:image_to_code}), we formalize our training data as image-caption-code triplets:
\begin{equation}
\mathcal{D}_{\text{train}} = \{(\mathbf{x}^{(i)}, \mathbf{t}_{\text{new}}^{(i)}, \mathbf{c}_{\text{new}}^{(i)})\}_{i=1}^{N}
\end{equation}
where $\mathbf{x}^{(i)}$ denotes our generated STEM image, $\mathbf{t}_{\text{new}}^{(i)}$ represents the code-grounded accurate caption, and $\mathbf{c}_{\text{new}}^{(i)}$ signifies the explanatory reproduction code. These three modalities constitute semantically equivalent representations of the same underlying STEM concept, providing complementary supervision signals for enhanced perceptual training.
We leverage these triplets to train our model through a two-stage paradigm: supervised finetuning followed by reinforcement learning.

\noindent \textbf{\emph{Stage 1: Supervised Finetuning (CodePercept-S1)}}
Using the Qwen3-VL series~\cite{Qwen3VL_github} as our base architecture, we jointly optimize two tasks: image captioning $(\mathbf{x}^{(i)}, \mathbf{t}_{\text{new}}^{(i)})$ and image-to-code translation $(\mathbf{x}^{(i)}, \mathbf{c}_{\text{new}}^{(i)})$. This joint training strategy provides benefits beyond training each task separately, as natural language captions $\mathbf{t}_{\text{new}}^{(i)}$ help the model understand visual content semantically, building strong visual understanding before generating code. Simultaneously, the explanatory code $\mathbf{c}_{\text{new}}^{(i)}$ acts as a structured caption that encodes visual information through executable programming constructs, providing precise spatial and quantitative details that complement natural language descriptions.

By interleaving these supervision signals, our S1 model learns to bridge visual perception, linguistic understanding, and symbolic code generation within a unified representation space. This allows captions to provide semantic context for code generation, while code generation improves caption accuracy through its precise and executable nature.

\noindent \textbf{\emph{Stage 2: Reinforcement Learning (CodePercept-R1)}}
We strategically apply reinforcement learning exclusively to code generation, as creating accurate, executable code is more challenging than natural language. Code requires strict syntax and logic, where minor errors can cause failures. Importantly, it provides inherently verifiable reward signals through executability and similarity metrics, making it well-suited for RL optimization. Building upon the SFT-trained model, we employ GRPO~\cite{guo2025deepseek,shao2024deepseekmath} with two defined rewards (Fig.\ref{fig:codepercept}) to further improve code quality:
\begin{itemize}[label={--}]
    \item \textbf{Format Reward} ($r_{\text{fmt}}$): A regular expression is used to validate that the generated code is encapsulated within a \texttt{```python ```} block, assigning a binary reward of 1.0 for a valid format and 0.0 otherwise.
    \item \textbf{Content Reward} ($r_{\text{cnt}}$): (1) \emph{Execution reward} $r_{\text{exec}}$ assigns 1 if the code executes successfully, otherwise 0; (2) \emph{Code-level Reward} ($r_{\text{code}}$):  A formatted score from GPT-4o~\cite{hurst2024gpt} that assesses the semantic equivalence between the generated code and the ground-truth code $\mathbf{c}$; (3) \emph{Image-level Reward} ($r_{\text{image}}$): A formatted score from GPT-4o that evaluates the visual similarity between the original image $\mathbf{x}$ and the rendered output (only available when code executes succeeded, otherwise 0). 
\end{itemize}
The overall reward is therefore defined as: $r = r_{\text{fmt}} + r_{\text{cnt}}$. Consider the standard GRPO approach, it samples a group of generated output set $\{o_{1}, o_{2},...,o_{G}\}$ for each query $q\in\mathcal{D}_{\rm train}$ (including question and image) from policy model $\pi_{\theta_{old}}$. Then GRPO maximizes the following objective and optimizes the model $\pi_{\theta}$:
\begin{align*}
\mathcal{J}(\theta) = \mathbb{E}_{\{o_{i}\}_{i=1}^{G} \sim \pi_{\theta_{\text{old}}}(O|q)} \Bigg[\frac{1}{G}\sum_{i=1}^{G} \min \Bigg( \frac{\pi_\theta(o_{i}|q)}{\pi_{\theta_{\text{old}}}(o_{i}|q)} A_{i}, \\
\text{clip}\left(\frac{\pi_\theta(o_{i}|q)}{\pi_{\theta_{\text{old}}}(o_{i}|q)}, 1-\epsilon, 1+\epsilon\right) A_{i} \Bigg) - \beta D_{KL}(\pi_\theta || \pi_{\text{ref}}) \Bigg]
\end{align*}
where $\epsilon$ and $\beta$ are the PPO clipping hyper-parameter
and the coefficient controlling the Kullback–Leibler (KL) penalty~\cite{guo2025deepseek,schulman2017proximal}, respectively. Specifically, for a group of $G$ outputs $\{o_1, \ldots, o_G\}$ sampled from the same input $q$, the advantage is calculated as:
\begin{align}
    A_i = \frac{r_i - \text{mean}({r_1, r_2, ..., r_G})}{\text{std}({r_1, r_2, ..., r_G})}
\end{align}
Through this reinforcement learning stage, the model is encouraged to generate not only syntactically correct and executable code, but also semantically accurate code that faithfully reconstructs the visual content.



\section{STEM2Code-Eval Benchmark}
\label{sec:benchmark}
\noindent \textbf{Why STEM2Code-Eval?}  Existing benchmarks~\cite{MathVista,MathVerse,MathVision} evaluate MLLMs through end-task problem-solving accuracy, which combine perceptual understanding with reasoning capabilities in STEM. When models fail, we cannot determine whether the failure stems from perceptual deficiencies or reasoning limitations. While a recent work~\cite{lu2025omnicaptioner} adopt a two-stage evaluation paradigm (image captioning followed by LLM solving) to isolate the perception capabilities of MLLMs, yet this metric only reflects the model's capacity to understand problem-relevant information rather than comprehensive visual perception. To close the gap, we propose a deterministic and verifiable paradigm, which requires models generate executable Python code that faithfully reproduces the original image. Only through complete and accurate visual comprehension can a model successfully reproduce the original image with high fidelity.

\noindent \textbf{What is STEM2Code-Eval?}

\noindent \emph{1) Source Image Collection.} We collect STEM images from test sets of six VQA-based STEM benchmarks~\cite{DynaMath,LogicVista,MathVerse,MathVision,MathVista,WeMath}, which span multiple STEM domains. 

\noindent \emph{2) Code Generation.} We first prompt Gemini2.5-Pro~\cite{comanici2025gemini} to generate detailed image captions about visual elements, spatial relationships, numerical values, \emph{etc}. Claude-Opus 4.1~\cite{anthropic2024claude} then combines the caption and image to synthesize reproducible Python code using \texttt{matplotlib} libraries.

\noindent \emph{3) Code Agent.} The generated code enters an iterative refinement loop: (i) rendering images via Python; (ii) applying LLM-based code repair upon execution failure; (iii) if successfully rendered, using Gemini2.5-Pro to score image similarity (style, content, functionality) against the original image and provide improvement suggestions, then refining the code to enhance reconstruction quality. This continues until the score exceeds 90\% or reaches 10 iterations.

\noindent \emph{4) Candidate Selection.} We rank all image-code pairs by similarity scores (reflecting reconstruction quality) and iteration counts (reflecting example difficulty), selecting the top 3k that score high on both dimensions for verification.

\noindent\emph{5) Visual Judge.} 
Ten expert annotators score candidates on a 5-point scale across three dimensions (style, content, functionality). The top 1,000 examples by average score are selected and undergo minor code refinement by human annotators to achieve pixel-perfect reproduction.
\section{Experiments}


\noindent \textbf{Implement Details}
We use the latest Qwen3-VL~\cite{Qwen3VL_github} as the base model. SFT is
trained on ICC-1M dataset for 1 epoch with SWIFT~\cite{zhao2025swift} on 32 A100 GPUs. The subsequent RL stage uses
VeRL~\cite{sheng2025hybridflow} for 1 epoch on 10k samples selected from our ICC-1M with the same hardware. Additional details about dataset curation, training settings, prompts, and visualizations are provided in supplementary materials.

\begin{table}[t]
    \centering
    \caption{Performance evaluation on our STEM2Code-Eval with 1k samples. We employ three metrics to assess their performances : (1) \textbf{Image Scoring} measures the visual similarity between the generated and original images; (2) \textbf{Code Scoring} assesses the quality, structure, and correctness of the generated Python code itself; (3) \textbf{Exec Rate} reports the execution success rate.
    }
    \vspace{-1em}
    \scalebox{0.8}{
    \begin{tabular}{l@{\hspace{6pt}}!{\vrule width 1.2pt}c@{\hspace{6pt}}c@{\hspace{6pt}}c@{\hspace{6pt}}!{\vrule width 1.2pt}@{\hspace{6pt}}c@{\hspace{6pt}}}
    \toprule[1.2pt]
    \textbf{Model} & \makecell[c]{\textbf{Image}\\\textbf{Score}} & \makecell[c]{\textbf{Code}\\\textbf{Score}} & \textbf{Avg.} & \makecell[c]{\textbf{Exec}\\\textbf{Rate}}\\
        \midrule
        Gemini2.5-Flash-Thinking & 57.25 & 60.87 &59.06 & 85.20  \\
        Claude-Opus 4.1-Thinking & 55.90 & 56.19 &56.05 & 97.10 \\
        GPT5-Thinking & 64.97 & 64.98 &64.98 & 96.60 \\
        Gemini2.5-Pro-Thinking & 68.89 & 75.41 &72.15 & 91.70 \\
        \midrule
        Intern-S1-8B~\cite{bai2025intern} &6.02&15.87&10.95 &26.60\\
        InternVL3.5-8B~\cite{wang2025internvl3} &13.50&18.16&15.83 &56.50\\
        MiniCPM-V-4.5~\cite{team2025minicpm4} &13.91&23.69&18.80 &50.80\\
        MiMo-VL-7B-RL~\cite{xiaomi2025mimo} &14.54&22.41&18.48 &60.30\\
        Ovis2.5-9B~\cite{lu2025ovis2} &9.76&11.26&10.51 &89.40\\
        KeyeVL1.5-8B~\cite{team2025kwai} &20.33&22.47&21.40 &73.40\\
        GLM-4.1V-9B~\cite{hong2025glm} &21.19&26.51&23.85 &72.00\\
        Qwen3-VL-4B-Thinking~\cite{Qwen3VL_github} & 25.38 & 34.53 &29.96 & 75.70 \\
        Qwen2.5-VL-72B-Instruct~\cite{bai2025qwen2} &32.82&25.83&29.33 &86.30\\
        Qwen3-VL-8B-Thinking~\cite{Qwen3VL_github} & 29.82 & 41.71 &35.77 & 78.90  \\
        Qwen3-VL-30A3B-Instruct~\cite{Qwen3VL_github} &33.05&31.04&32.05 &87.50\\
        Seed1.6-Vision-nothinking~\cite{guo2025seed1} & 31.22 & 38.56 &34.89 & 85.50 \\
        Qwen3-VL-32B-Instruct~\cite{Qwen3VL_github} &36.85&39.98&38.42 &81.80\\
        Qwen3-VL-30A3B-Thinking~\cite{Qwen3VL_github} &37.47&35.53&36.50 &87.10\\
        Qwen3-VL-Plus-Instruct~\cite{Qwen3VL_github} & 45.94 & 40.40 &43.17 & 90.00 \\
        Qwen3-VL-Plus-Thinking~\cite{Qwen3VL_github} & 45.59 & 40.61 &43.10 & 89.20 \\
        Seed1.6-Vision-Thinking~\cite{guo2025seed1} & 42.03 & 40.74 &41.39 & 94.70 \\
        \midrule
        Qwen3-VL-4B-Instruct~\cite{Qwen3VL_github} & 24.55 & 26.42 &25.49 & 79.40 \\
        \rowcolor{cyan!5} \bfseries{CodePercept-4B-S1} & 38.13 & 43.43 &40.78 & 80.70\\
        \rowcolor{cyan!10} \bfseries{CodePercept-4B-R1} & \textbf{47.17} & \textbf{45.86} &\bfseries{46.52} & \bfseries{91.30}\\
        \midrule
        Qwen3-VL-8B-Instruct~\cite{Qwen3VL_github} & 28.59 & 28.23 &28.41 & 85.30 \\
        \rowcolor{cyan!5} \bfseries{CodePercept-8B-S1} & 44.53 & 46.78 &45.66 & 87.60 \\
        \rowcolor{cyan!10} \bfseries{CodePercept-8B-R1} & \textbf{50.25} & \textbf{47.04} &\bfseries{48.65} & \bfseries{93.40} \\
        \midrule
        Qwen3-VL-32B-Instruct~\cite{Qwen3VL_github} &36.85&39.98&38.42 &81.80\\
        \rowcolor{cyan!5} \bfseries{CodePercept-32B-S1} & 61.14 & 56.99 &59.07 & 93.00 \\
        \rowcolor{cyan!10} \bfseries{CodePercept-32B-R1} &\bfseries{68.97}  &\bfseries{62.53}  &\bfseries{65.75} &\bfseries{95.90}  \\
        \bottomrule
    \end{tabular}
    }
    \label{STEM2code}
    \vspace{-1em}
\end{table}

\begin{table*}[t]
    \centering
    \setlength{\tabcolsep}{3.5pt}
    \caption{Ablation study on perception ability, evaluated using the captioner-solver setup. More accurate descriptions are expected to yield better reasoning results. ``IM", ``ID", ``SG" refer to Image Reproduce, Image Diversity, and Solid Geometry pipeline, respectively. ``CodeCap" and ``ImCode" refer to code-grounded captions and explanatory image-code pairs, respectively.}
    \vspace{-1em}
    \label{tab:ablations}
    \sisetup{
        table-format=2.2, 
        detect-weight,    
        mode=text         
    }
    \scalebox{0.9}{
    \begin{tabular}{@{} l l SSSSSS S @{}}
        \toprule
        \multirow{2}{*}{Group} & \multirow{2}{*}{\textbf{Image Captioner}} & \multicolumn{6}{c}{Benchmark Datasets (\%)} & {\multirow{2}{*}{Average}} \\
        \cmidrule(lr){3-8}
        & & {MathVision} & {MathVista} & {MathVerse} & {DynaMath} & {WeMath} & {LogicVista} & \\
        
        \midrule
        \multicolumn{8}{c}{\emph{\textbf{LLM Solver: Qwen3-30A3-Thinking~\cite{yang2025qwen3}}}} \\
        \midrule
        \multirow{4}{*}{1} & Qwen3-VL-8B-Instruct~\cite{Qwen3VL_github} &54.37  & 69.60 & 63.75 & 72.19 & 45.43 & 56.82 & 60.36 \\
        & \quad + IR-CodeCap &55.86  &69.50  &63.50  &72.59  &46.95  &57.05   &60.91 \\
        & \quad + ID-CodeCap &58.32  &71.20  &64.52  &71.70  &46.76  &60.40   &62.15 \\
        & \quad + SG-CodeCap &59.93  &68.81  &66.54  &72.73  &47.05  &62.64   &62.75 \\
        \midrule
        \multirow{2}{*}{2} & NativeCap &56.20  &68.39  &63.87  &70.69  &46.48  &59.06   &60.78 \\
        & CodeCap &59.93  &68.81  &66.54  &72.73  &47.05  &62.64   &62.75 \\
        \midrule
        \multirow{2}{*}{3} & CodeCap &59.93  &68.81  &66.54  &72.73  &47.05  &62.64  &62.75 \\
        & CodeCap + ImCode &59.31  &70.20  &66.52  &73.20  &49.14  &61.52  &63.32 \\
        \bottomrule
    \end{tabular}
    }
\end{table*}
\subsection{Main Results}

\noindent \textbf{1) Problem-solving Perception Evaluation}
We evaluate our CodePercept model against previous SOTA MLLMs in their capacity as image captioners, subsequently employing the same LLM (Qwen3-30A3-Thinking and Qwen3-235A22-Thinking)~\cite{yang2025qwen3} as the solver to generate final answers across six widely-adopted STEM reasoning benchmarks. More accurate image descriptions should yield better reasoning results.
Tab.~\ref{two-stage math} presents comprehensive comparisons across all benchmarks, where our CodePercept model demonstrate consistent and substantial improvements over baseline MLLMs. Under the Qwen3-30A3-Thinking solver, CodePercept-4B-S1 outperforms Qwen3-VL-4B-Instruct by 2.8\%. When scaled to 8B parameters, the gains further increase to 3\%. When leveraging the more powerful Qwen3-235A22-Thinking solver, CodePercept-4B/8B-S1 maintains comparable improvements (2.9\% and 3.4\%, respectively), demonstrating remarkable robustness across different solver capabilities. Notably, CodePercept-8B-S1 surpasses several substantially larger models, including Qwen2.5-VL-72B (by 6.2\%), and even approaches the performance of frontier models like Claude-Opus 4.1-Thinking and GPT5-Thinking. These results validate our central hypothesis: code-grounded perception serves as an effective paradigm for enhancing STEM visual perception.

\noindent \textbf{2) Image Reproduce Perception Evaluation}
Problem-solving metric only reflects the model’s capacity to understand problem-relevant information rather than comprehensive visual perception.
To directly assess visual perception, we introduce STEM2Code-Eval to support image reproduce perception evaluation. Tab.~\ref{STEM2code} presents comprehensive results. Through joint training two code-grounded tasks, CodePercept-S1 models achieve substantial improvements, with CodePercept-4B-S1 reaching 54.09 (+10.6 over Qwen3-VL-4B-Instruct) and CodePercept-8B-S1 achieving 59.64 (+12.3 over Qwen3-VL-8B-Instruct). These gains validate grounding perception in executable code fundamentally enhances strong visual perception.

We further optimize code generation using RL to enforce executability and reconstruction accuracy. CodePercept-4B-R1 achieves 61.44 (+7.35) and CodePercept-8B-R1 reaches 63.56 (+3.92). Both models surpass super-large models including Seed1.6-Vision and Qwen3-VL-Plus by 2.29-2.97 and 4.41-5.09, respectively. 

\subsection{Ablation Study}

\noindent \textbf{1) Three Pipeline Comparison} We evaluate three data generation strategies by training models on captions generated via Code-Grounded Caption Generation (Sec.~\ref{subsec:code_grounded_caption}). Tab.~\ref{tab:ablations} (Group 1) shows a gradual increase in the average score (60.91 $\to$ 62.15 $\to$ 62.75), demonstrating the effectiveness of our three pipelines. 
In particular, ID-CodeCap, which includes diverse, procedurally-generated STEM images with verified code provide stronger training signals, yields the greatest performance gain.

\noindent \textbf{2) Code-Grounded Caption Generation} To validate our code-grounded approach (Sec.~\ref{subsec:code_grounded_caption}), we compare it against direct caption generation (NativeCap), where we prompt Gemini2.5-pro~\cite{comanici2025gemini} with identical instructions to generate image captions directly without code analysis on the same ICC-1M data in Group 2. Our analysis yields two key insights. First, even NativeCap improves the average score to 60.36 over the 60.78 baseline, demonstrating the general effectiveness of our data pipeline. Second and more importantly, our code-grounded method (CodeCap) further boosts performance to 62.75, achieving a significant 2.0 gain over NativeCap. This validates our hypothesis that by extracting visual facts from executable code, CodeCap effectively reduces hallucinations in numerical values and geometric details common in direct vision-language generation.

\noindent \textbf{3) STEM Image-to-Code Translation}
Adding ImCode data (Group 3) further boosts performance to 63.32, gaining 0.6 over CodeCap alone. This shows that code serves as a complementary modality for image caption. Image-to-caption and image-to-code tasks reinforce each other: captions provide semantic context for code generation, while 
code overcomes the inherent limitations of natural language in describing complex mathematical visuals through its unique structured representation. Together, both tasks demonstrate that code is essential for reliable STEM image perception: whether as intermediate supervision for caption synthesis or as a direct target for structured understanding.

\section{Conclusion}
We identify visual perception as the primary bottleneck for MLLMs in STEM. To address this, we introduce \textbf{CodePercept}, a paradigm that leverages executable Python code to enhance perception. Our contributions include \textbf{ICC-1M}, a large-scale image-caption-code triplets, enabling two novel training approaches: Code-Grounded Caption Generation and STEM Image-to-Code Translation. We also develop \textbf{STEM2Code-Eval}, a benchmark that evaluates perception by generating executable code to reconstruct an image, beyond traditional problem-solving proxies. Experimental results prove the effectiveness of our CodePercept, establishing that executable codes, whether as intermediate supervision for caption synthesis or as a direct target for structured understanding, are essential for enhancing perception.
{
    \small
    \bibliographystyle{ieeenat_fullname}
    \bibliography{main}
}
\clearpage
\onecolumn
\setcounter{page}{1}
\makeMySupplementaryTitle
\appendix
\lstset{
    language=Python,
    basicstyle=\ttfamily\footnotesize,
    backgroundcolor=\color{codegray},
    commentstyle=\color{codegreen},
    keywordstyle=\color{blue},
    stringstyle=\color{codepurple},
    breaklines=true,
    breakatwhitespace=true,
    showstringspaces=false,
    numbers=left,
    numberstyle=\tiny\color{gray},
    frame=single,
    tabsize=4,
    captionpos=b,
    escapeinside={(*@}{@*)}, 
}
\newtcolorbox{takeawaybox}[1]{ 
  breakable, 
  title={#1}, 
  colback=cvprblue!10!white,
  colframe=cvprblue,
  fonttitle=\bfseries,
  toptitle=1mm,
  bottomtitle=1mm,
  pad before break=2mm,
  pad after break=2mm,
  title after break={{#1} (continued)}, 
}

\section{Unified Quality Control}
\label{supp:Unified_Quality_Control}
To ensure the reliability and consistency of our data generation pipeline, we implement a unified quality control framework with explicit prompts for each stage. Our quality control operates at three critical checkpoints: (1) verifying the correctness and quality of generated code, (2) verifying the correctness and quality of rendered image, and (3) verifying the consistency of image-code pairs. Below, we present the detailed prompts used at each stage to maintain high standards of both technical accuracy and instructional effectiveness.

\begin{takeawaybox}{Code Quality Prompt $\mathcal{Q}_{\rm C}$}
  \small
  \textit{You are a leading AI scientist tasked with curating a training dataset for an Image-to-Code model designed to understand and generate expert-level human code. Your role combines philosophical insight with engineering rigor as a quality assessor.}
  \begin{enumerate}[label=\textbf{\arabic*.}, itemsep=8pt, leftmargin=*]
    \item \textit{
      \textbf{Core Philosophy: Maximize Semantic Value}
      \newline
      \textbf{Objective}: Cultivate the model's ability to map visual patterns to high-level programming concepts by distinguishing meaningful abstractions from arbitrary procedures.
      \newline
      \textbf{1.1 Conceptual Framework}
      \begin{enumerate}[label=\arabic*., itemsep=4pt, leftmargin=*]
        \item \textbf{High-Value Abstraction} (samples we pursue):
          \begin{itemize}
            \item Code implements mathematical concepts or algorithms with clear, visually recognizable structural features
            \item Procedural steps are necessary components for realizing the abstraction
            \item Examples: $y = x^2$ (parabola), Fibonacci spiral algorithms, \texttt{noise.pnoise2} (procedural noise)
          \end{itemize}
        \item \textbf{Low-Value Procedure} (samples we reject):
          \begin{itemize}
            \item Code contains arbitrary, context-free computational steps not inferable from the final image
            \item Contains ``magic numbers" and ``black-box operations" toxic to model learning
            \item Examples: \texttt{data += 100}, \texttt{data = data * 1.1}, \texttt{if unseen\_variable: ...}
          \end{itemize}
      \end{enumerate}
    }
    \item \textit{
      \textbf{Evaluation Rules}
      \newline
      \textbf{2.1 Pattern Recognition Rule}
      \begin{enumerate}[label=\arabic*., itemsep=4pt, leftmargin=*]
        \item \textbf{[Qualified]} Image exhibits strong, non-random structural patterns:
          \begin{itemize}
            \item Symmetry, spirals, fractals, waveforms, regular grids
            \item Code clearly implements mathematical formulas or algorithms producing the pattern
            \item This represents the highest-value sample type
          \end{itemize}
        \item \textbf{[Qualified]} Image appears random but uses domain-standard tools:
          \begin{itemize}
            \item Examples: noise maps generated via \texttt{noise.pnoise2}
            \item Teaches proper use of professional libraries
          \end{itemize}
        \item \textbf{[Qualified]} Simple graphics with hardcoded constants:
          \begin{itemize}
            \item Bar charts, scatter plots with direct data declaration (\texttt{data = [1,2,3]})
            \item Basic but acceptable ``declarative" samples
          \end{itemize}
        \item \textbf{[Disqualified]} True random generation without structure:
          \begin{itemize}
            \item Generated by \texttt{np.random.rand()} with no recognizable macroscopic patterns
          \end{itemize}
      \end{enumerate}
      \textbf{2.2 Computational Intent Rule}
      \begin{enumerate}[label=\arabic*., itemsep=4pt, leftmargin=*]
        \item \textbf{[Qualified]} Computational steps are intrinsic to high-value abstraction:
          \begin{itemize}
            \item Example: Z-value-based color/linewidth computation in torus knot code implementing ``pseudo-3D depth perception" (a universal rendering technique)
          \end{itemize}
        \item \textbf{[Disqualified]} Arbitrary, isolated numerical operations:
          \begin{itemize}
            \item Logic or context completely lost in the image
            \item Example: \texttt{aqi\_data[0, :] += 100}
          \end{itemize}
      \end{enumerate}
      \textbf{2.3 Dependency \& Source Rule}
      \begin{enumerate}[label=\arabic*., itemsep=4pt, leftmargin=*]
        \item \textbf{[Qualified]} Dependencies limited to:
          \begin{itemize}
            \item Python standard library
            \item Well-known pip-installable libraries directly related to visual generation (\texttt{numpy}, \texttt{scipy}, \texttt{noise}, etc.)
          \end{itemize}
        \item \textbf{[Disqualified]} Reliance on inaccessible external resources:
          \begin{itemize}
            \item External files (.csv, .json), databases, or network APIs
            \item Model cannot access these resources during inference
          \end{itemize}
      \end{enumerate}
    }
    \item \textit{
      \textbf{Required Output Format}
      \begin{itemize}
        \item \textbf{[Verdict]}: (Qualified / Disqualified)
        \item \textbf{[Suitability Score]}: (1-5, where 1=harmful, 3=baseline qualified, 5=exceptionally educational)
        \item \textbf{[Rationale]}: Briefly explain the verdict. Explicitly identify whether the code represents \textbf{[High-Value Abstraction]} or \textbf{[Low-Value Procedure]}. If qualified, articulate the ``core concept" being taught (e.g., ``recognizing Fibonacci spiral algorithms" or ``learning pseudo-3D rendering techniques"). If disqualified, specify which computational step is ``arbitrary and non-invertible."
      \end{itemize}
      Evaluate the provided Python visualization code according to these principles.
    }
  \end{enumerate}
\end{takeawaybox}

\begin{takeawaybox}{Image Quality Prompt $\mathcal{Q}_{\rm I}$}
  \small
  \textit{You are a professional image quality analyst. Your mission is to conduct a quality review of a given code-generated image according to the strict standards defined below. Your report must not only include the final verdict but, more importantly, \textbf{must demonstrate a clear, systematic analysis process} to prove that your conclusion is well-considered.}
  \begin{enumerate}[label=\textbf{\arabic*.}, itemsep=8pt, leftmargin=*]
    \item \textit{
      \textbf{Stage 1: Technical Rendering Audit}
      \newline
      \textbf{Task}: Assess whether the image has fatal technical rendering defects.
      \newline
      \textbf{Error Classification Definitions}
      \begin{enumerate}[label=\textbf{\Alph*.} (Class), itemsep=4pt, leftmargin=*]
        \item \textbf{Fatal Rendering Errors}: Technical rendering failures caused by algorithmic or data interpretation failures that \textbf{severely compromise} the geometry, paths, or annotations of the graphic.
          \begin{itemize}
            \item \textbf{A-1. Vector Path Catastrophe}:
                \begin{itemize}
                    \item \textit{Path Distortion}: Paths exhibit \textbf{severe} unexpected spikes, kinks, or non-smoothness.
                    \item \textit{Projection Artifacts}: A portion of a path or fill is \textbf{severely stretched} to or beyond the canvas boundary.
                    \item \textit{Unexpected Unclosed Path}: A shape that should be closed has a \textbf{significant gap}, causing the fill color to ``leak".
                \end{itemize}
            \item \textbf{A-2. Geometric Annotation Error}:
                \begin{itemize}
                    \item \textit{Severe Anchor Misplacement}: The anchor point of an annotation element (e.g., angle arc, dimension arrow) is \textbf{completely detached} from the object it is supposed to annotate.
                \end{itemize}
            \item \textbf{A-3. Empty Chart}:
                \begin{itemize}
                    \item The coordinate system and labels are drawn correctly, but the \textbf{data series} (e.g., lines, bars) is \textbf{entirely missing}.
                \end{itemize}
            \item \textbf{A-4. Indiscernible Density}:
                \begin{itemize}
                    \item Contains a large number of elements that are indescribable, chaotic, indistinguishable, densely overlapping, or blurry (e.g., a ``point cloud soup" or a ``sea of lines").
                \end{itemize}
            \item \textbf{A-5. Layout Anomaly}:
                \begin{itemize}
                    \item Abnormal layout resulting in large areas of whitespace.
                    \item Crowded layout making key elements difficult to distinguish.
                \end{itemize}
          \end{itemize}
        \item \textbf{Non-Fatal Design Flaws}: The technical rendering is successful, but there are flaws in aesthetics or content logic.
          \begin{itemize}
            \item \textit{Examples}: Poor color choices, normal occlusion between elements, unintuitive logical layer order, questionable content that is drawn correctly.
          \end{itemize}
      \end{enumerate}
    }
    \item \textit{
      \textbf{Core Principles \& Output Format}
      \begin{itemize}
        \item \textbf{Principle 1: Identify Class A Errors (Critical)}. If \textbf{at least one significant, unintentional, technical Class A rendering error} is present, the verdict is \textbf{Fail}.
        \item \textbf{Principle 2: Exempt All Minor Issues}. \textbf{Extremely minor technical glitches} (e.g., single-pixel offsets) and \textbf{all Class B design flaws} (e.g., aesthetic issues), regardless of their number, \textbf{must never} be the basis for a ``Fail" verdict.
      \end{itemize}
      \textbf{Output Format}: Strictly follow the process above, analyzing step-by-step and providing reasons. Then, generate the complete analysis report using the specified Markdown format below.
      \newline \newline
      \texttt{\#\# Rendering Quality Audit} \newline
      \texttt{* Final Verdict: [Enter ``Pass" or ``Fail"]} \newline \newline
      \texttt{\#\#\# Error Analysis} \newline
      \texttt{[Note: This section should only be generated if the final verdict is ``Fail". If multiple errors exist, repeat the list item.]} \newline
      \texttt{* Error Type: [Enter A-1, A-2, etc.]} \newline
      \texttt{* Description: [Describe the specific error phenomenon and its location in the image.]}
    }
  \end{enumerate}
\end{takeawaybox}

\begin{takeawaybox}{Image-Code Consistency Prompt $\mathcal{Q}_{\rm IC}$}
  \small
  \textit{You are a professional image quality analyst. Your mission is to conduct a quality review of the given code-image pair according to the strict standards defined below.}
  \newline
  \textbf{Task}: Determine if the \textbf{core structure and scene} presented in the image are fundamentally consistent with the \textbf{design intent} of the code.

  \begin{enumerate}[label=\textbf{\arabic*.}, itemsep=8pt, leftmargin=*]
    \item \textit{
      \textbf{Workflow}
      \begin{enumerate}[label=\arabic*., itemsep=4pt, leftmargin=*]
        \item \textbf{Visual Evidence Inventory}: \textbf{Ignore the code} and visually inventory the elements in the image (quantity, color, layout, relationships).
        \item \textbf{Code Logic Deduction (Design Blueprint)}: \textbf{Mentally compile the code}, analyzing variables (e.g., \texttt{n=8}), loops (e.g., \texttt{for i in range(8)}), and control flow to precisely deduce the image's \textbf{quantitative specifications} (e.g., total number of elements, exact coordinates, connection relationships, etc.).
        \item \textbf{Critical Comparison}: Strictly compare the ``Visual Evidence" with the ``Design Blueprint".
      \end{enumerate}
    }
    \item \textit{
      \textbf{Core Verdict Standard: ``Fundamental Mismatch"}
      \begin{itemize}[itemsep=4pt, leftmargin=*]
        \item \textbf{``Sufficient Match"}: The \textbf{core elements, deduced quantities, attributes, and spatial/structural relationships} described by the code are fundamentally consistent with the image. Minor rendering differences (e.g., color shades, anti-aliasing effects) are permissible.
        \item \textbf{``Fundamental Mismatch"}: There is a \textbf{structural or categorical} essential difference between the code's intent and the image's reality. Any of the following conditions qualify:
          \begin{itemize}[itemsep=2pt, leftmargin=15pt]
            \item[\textbf{a.}] \textbf{Structural/Layout/Relational Breakdown}: The organization of elements or their spatial relationships have completely collapsed. \newline \textit{Example: Code intends to generate a geometric puzzle of \textbf{neatly arranged} shapes, but the image shows them \textbf{randomly piled up with misaligned vertices}.}
            \item[\textbf{b.}] \textbf{Categorical Difference}: The core scene or element type is entirely different. \newline \textit{Example: Code draws a bar chart, but the image is a scatter plot.}
            \item[\textbf{c.}] \textbf{Severe Mismatch of Key Elements/Attributes}: The \textbf{quantity}, type, or core attributes of key elements are severely missing or incorrect. \newline \textit{Example: Code intends to draw a 10-pointed star, but the image shows only 3 unrelated lines.}
            \item[\textbf{d.}] \textbf{Structural Collapse or Loss of Integrity}: The elements in the image are fragmented or incomplete, failing to form the complete structure intended by the code. \newline \textit{Example: Code intends to draw a \textbf{complete, closed ellipse}, but the image only shows two \textbf{separate, unclosed curves}.}
            \item[\textbf{e.}] \textbf{Data-Level Misalignment}: The data order, sorting, or mapping relationship between elements does not match the code's intent. \newline \textit{Example: Code intends to connect points in the order \texttt{A->B->C}, but the image shows \texttt{A->C->B}.}
          \end{itemize}
      \end{itemize}
    }
    \item \textit{
      \textbf{Instructions}
      \newline
      Now, here is the code used to generate the image: \{input code this\}
      \newline
      Strictly follow the process above, analyze step-by-step, provide reasons, and then generate the complete analysis report using the specified Markdown format below.
    }
    \item \textit{
        \textbf{Output Format}
        \newline
        \texttt{\# Image Quality Analysis Report} \newline
        \texttt{\#\# Consistency Review} \newline
        \texttt{* Verdict: [Enter ``Sufficient Match" or ``Fundamental Mismatch"]} \newline
        \texttt{* Reason: [Fill this in only if the verdict is ``Fundamental Mismatch", explaining the type and specific phenomenon of the mismatch.]}
    }
  \end{enumerate}
\end{takeawaybox}

\section{Native Image-Code Pairs \emph{v.s.} Explanatory Image-Code Pairs}
\label{supp:Explanatory_Image_Code_Pairs}
\subsection*{1 The Definition of Native Image-Code Pairs}
Native image-code pairs consist of raw visual image paired with ground-truth executable code without additional explanatory context. While these pairs ensure code correctness, they lack the pedagogical scaffolding necessary for models to learn the visual-to-code mappings and the rationale for specific implementation choices.

\subsection*{2 The Definition of Explanatory Image-Code Pairs}
Explanatory image-code pairs augment ground-truth code with step-by-step descriptions, explicit parameter justifications, and instructional commentary that articulates how visual elements map to code constructs. This explanatory richness enables models to understand not merely what code to generate, but why particular implementations best capture the visual content.

To illustrate the enhanced instructional value of explanatory image-code pairs, we present a representative example below. The example demonstrates how the explanatory framework transforms bare code into comprehensive learning materials by integrating visual observation, implementation logic, and detailed commentary.

\begin{takeawaybox}{A Explanatory Image-Code Pair Example}
  \small
  \begin{center}
    \includegraphics[width=0.2\textwidth]{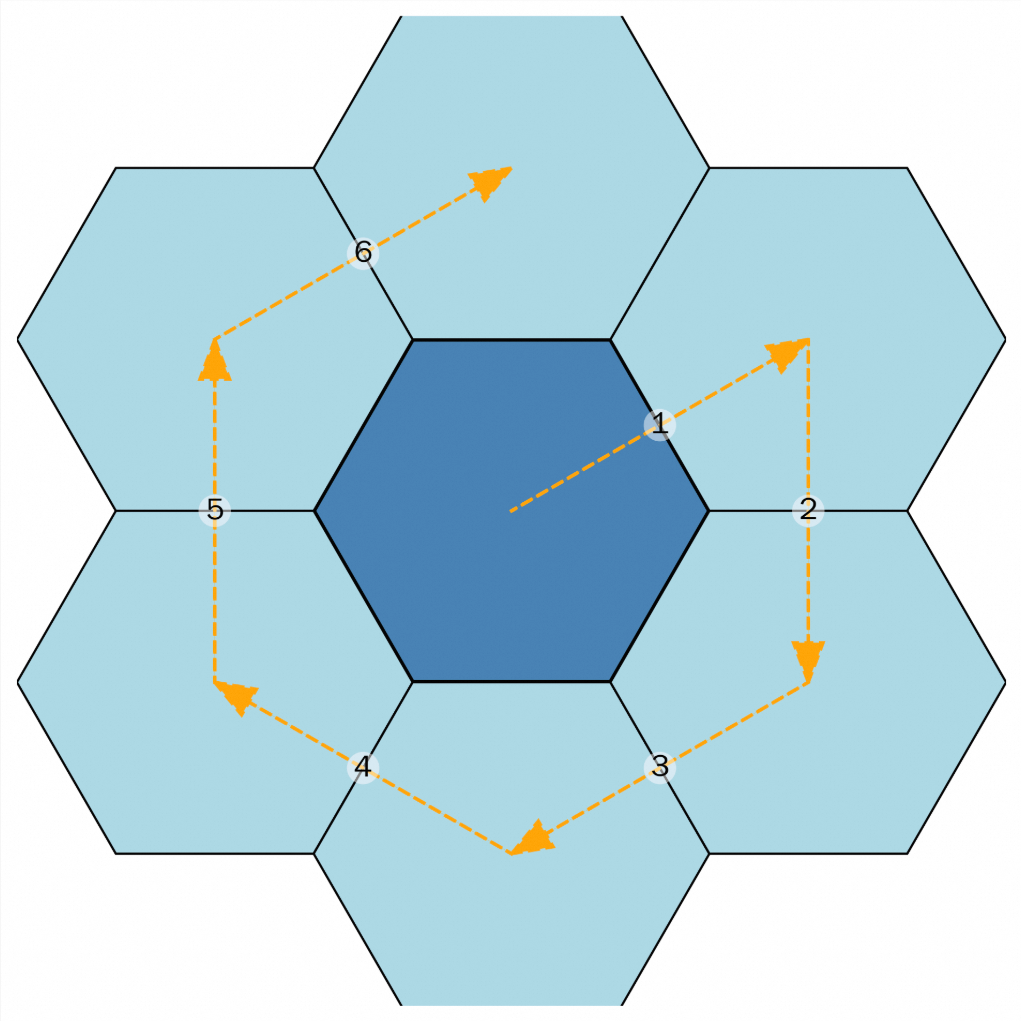}
    \captionof{figure}{The hexagonal grid with a spiraling path, generated by the Python code below.}
    \label{fig:hexagon_path}
  \end{center}
  \textit{An analysis of the image reveals a honeycomb structure composed of seven regular hexagons. A central, darker blue hexagon is surrounded by six lighter blue hexagons. A dashed orange path originates inside the central hexagon and spirals outwards, passing through the vertices of the central hexagon. Below is a breakdown of the steps to reproduce this image using Python's matplotlib library, followed by the complete, executable code.}
  \begin{enumerate}[label=\textbf{\arabic*.}, itemsep=8pt, leftmargin=*]
    \item \textit{
      \textbf{Implementation Breakdown}
      \newline
      To reproduce this, we can use the \texttt{matplotlib} library in Python. Here's a breakdown of the steps:
      \begin{enumerate}[label=\arabic*., itemsep=4pt, leftmargin=*]
        \item \textbf{Hexagon Grid}: We define a function to calculate the vertices of a regular ``flat-top" hexagon given its center and size. The central hexagon is placed at the origin $(0, 0)$, and the six surrounding hexagons are arranged in a standard hexagonal grid pattern around it.
        \item \textbf{Drawing Hexagons}: The \texttt{matplotlib.patches.Polygon} class is used to create the hexagon shapes. The central hexagon is colored \texttt{steelblue}, and the outer ones are \texttt{lightblue}. All have black edges, with the central one having a slightly thicker border.
        \item \textbf{Defining the Path}: The path is a complex polyline. It starts near the center, moves to the first vertex ($p_1$), then ``bounces" into the adjacent hexagon to an intermediate point ($mid_1$), and then connects to the next vertex ($p_2$). This pattern repeats for all six vertices of the central hexagon ($p_1$ through $p_6$).
        \item \textbf{Drawing the Path and Arrows}: The path is constructed from two types of segments:
          \begin{itemize}
            \item[$\circ$] Simple dashed lines, drawn with \texttt{ax.plot()}.
            \item[$\circ$] Dashed lines with an arrowhead, drawn with \texttt{ax.annotate()}. The annotate function is used here to draw an arrow from a start point (\texttt{xytext}) to an end point (\texttt{xy}) with specific styling (dashed, orange, with a particular arrowhead shape). This is used for the first leg of each ``bounce" (e.g., $p_1 \rightarrow mid_1$).
          \end{itemize}
        \item \textbf{Labels and Markers}: The vertices 1 through 6 are marked with a small white circle with a black outline, created using \texttt{ax.plot()} with an \texttt{'o'} marker. The corresponding numbers are placed nearby using \texttt{ax.text()}. The positions of the labels and the intermediate path points are carefully chosen to match the original image.
        \item \textbf{Final Touches}: The axes are turned off, and the plot limits are set to frame the hexagons similarly to the original image, creating a clean and focused visual.
      \end{enumerate}
    }
    \item \textit{\textbf{Python Implementation Code}
    \newline
      The following code reproduces the honeycomb structure:}
      
\begin{lstlisting}
import matplotlib.pyplot as plt
import numpy as np
from matplotlib.patches import Polygon

def generate_hexagon_path_image():
    """
    Generates the code to reproduce the image of 
    a hexagonal grid with a path.
    """
    # 1. Setup figure and axes
    fig, ax = plt.subplots(figsize=(10, 10))
    ax.set_aspect('equal')
    ax.axis('off')
    fig.set_facecolor('white')

    # 2. Parameters
    s = 1.0  # Hexagon side length
    dark_blue_color = 'steelblue'
    light_blue_color = '#add8e6'
    orange_color = 'orange'
    line_props = dict(color=orange_color, 
                      linestyle='--', 
                      linewidth=2, 
                      zorder=2)
    arrow_props = dict(
        arrowstyle='->, head_length=0.7, head_width=0.4',
        **line_props)

    # 3. Hexagon generation function (flat-top)
    def get_flat_top_hexagon_vertices(center, size):
        x, y = center
        return np.array([
            [x + size, y],
            [x + size / 2, y + size * np.sqrt(3) / 2],
            [x - size / 2, y + size * np.sqrt(3) / 2],
            [x - size, y],
            [x - size / 2, y - size * np.sqrt(3) / 2],
            [x + size / 2, y - size * np.sqrt(3) / 2],
        ])

    # 4. Draw hexagons
    center_main = np.array([0, 0])
    vertices_main = get_flat_top_hexagon_vertices(
        center_main, s)
    ax.add_patch(Polygon(vertices_main, 
                         facecolor=dark_blue_color, 
                         edgecolor='black', 
                         linewidth=2))

    centers_outer = [
        center_main + [0, s * np.sqrt(3)],
        center_main + [s * 1.5, s * np.sqrt(3) / 2],
        center_main + [s * 1.5, -s * np.sqrt(3) / 2],
        center_main + [0, -s * np.sqrt(3)],
        center_main + [-s * 1.5, -s * np.sqrt(3) / 2],
        center_main + [-s * 1.5, s * np.sqrt(3) / 2],
    ]
    for center in centers_outer:
        ax.add_patch(Polygon(
            get_flat_top_hexagon_vertices(center, s),
            facecolor=light_blue_color, 
            edgecolor='black', 
            linewidth=1))

    # 5. Define path vertices and intermediate points
    p2, p1, p6, p5, p4, p3 = vertices_main

    mid1 = np.array([1.05, 0.45])
    mid2 = np.array([1.05, -0.45])
    mid3 = np.array([0, -1.3])
    mid4 = np.array([-1.05, -0.45])
    mid5 = np.array([-1.05, 0.45])
    mid6 = np.array([-0.2, 1.4])
    
    path_start = np.array([-0.4, 0])
    path_final_end = np.array([-0.7, 1.8])

    # 6. Draw the path segments
    ax.plot([path_start[0], p1[0]], 
            [path_start[1], p1[1]], 
            **line_props)

    path_segments = [
        (p1, mid1, p2),
        (p2, mid2, p3),
        (p3, mid3, p4),
        (p4, mid4, p5),
        (p5, mid5, p6)
    ]

    for start_node, mid_node, end_node in path_segments:
        ax.annotate('', xy=mid_node, 
                    xytext=start_node, 
                    arrowprops=arrow_props)
        ax.plot([mid_node[0], end_node[0]], 
                [mid_node[1], end_node[1]], 
                **line_props)

    ax.annotate('', xy=mid6, xytext=p6, 
                arrowprops=arrow_props)
    ax.plot([mid6[0], path_final_end[0]], 
            [mid6[1], path_final_end[1]], 
            **line_props)

    # 7. Add vertex markers and labels
    vertex_map = {'1': p1, '2': p2, '3': p3, 
                  '4': p4, '5': p5, '6': p6}
    label_positions = {
        '1': (p1[0] - 0.25, p1[1] - 0.1),
        '2': (p2[0] + 0.15, p2[1] - 0.1),
        '3': (p3[0] + 0.15, p3[1] - 0.1),
        '4': (p4[0] - 0.25, p4[1] - 0.1),
        '5': (p5[0] - 0.2, p5[1]),
        '6': (p6[0] - 0.25, p6[1] - 0.1),
    }

    for num, pos in vertex_map.items():
        ax.plot(pos[0], pos[1], 'o', 
                markersize=6, 
                markerfacecolor='white', 
                markeredgecolor='black', 
                zorder=4)
        label_pos = label_positions[num]
        ax.text(label_pos[0], label_pos[1], num, 
                fontsize=14, 
                ha='center', 
                va='center', 
                zorder=5)

    # 8. Finalize plot
    ax.set_xlim(-2.8, 2.8)
    ax.set_ylim(-2.8, 2.8)

    plt.show()

if __name__ == '__main__':
    generate_hexagon_path_image()
\end{lstlisting}
    \item \textit{\textbf{Usage Instructions}
      \begin{itemize}[itemsep=4pt]
        \item Requires: matplotlib, numpy libraries
        \item Run the script to generate the visualization
        \item Adjust parameters in section 2 to customize appearance
        \item Modify \texttt{s} variable to change hexagon size
      \end{itemize}
      Note: This structured approach allows for faithful reproduction of hexagonal grid patterns with customizable paths.
      }
  \end{enumerate}
\end{takeawaybox}

\section{Native Caption \emph{v.s.} Code-Grounded Caption}
\label{supp:Code_Grounded_Caption}
\subsection*{Comparisons}
In this subsection, we present a representative example to illustrate the comparative advantages of our approach. The provided image poses a significant challenge: it contains an 8×8 grid with varying dot patterns in each cell, interconnected by red arrows that simulate the L-shaped movement of a knight in Chinese chess, traversing all cells to form a complex path. 

We observe that even state-of-the-art MLLMs, such as Gemini 2.5 Pro, produce detailed descriptions riddled with factual errors when analyzing this image—misidentifying node counts (e.g., claiming 1 node instead of 2 in cell (1,8)), describing incorrect spatial arrangements, and reporting 68 arrows instead of the actual 63. 

In contrast, our code-grounded captions deliver precise, verifiable information. By anchoring descriptions in executable code that can be validated against actual image data, this approach systematically eliminates perceptual hallucinations common in vision-language models, ensuring the blueprint-level accuracy essential for tasks demanding exact spatial reasoning and faithful reconstruction.

\begin{takeawaybox}{Comparisons on Native Caption and Code-Grounded Caption}
  \begin{center}
    \includegraphics[width=0.4\textwidth]{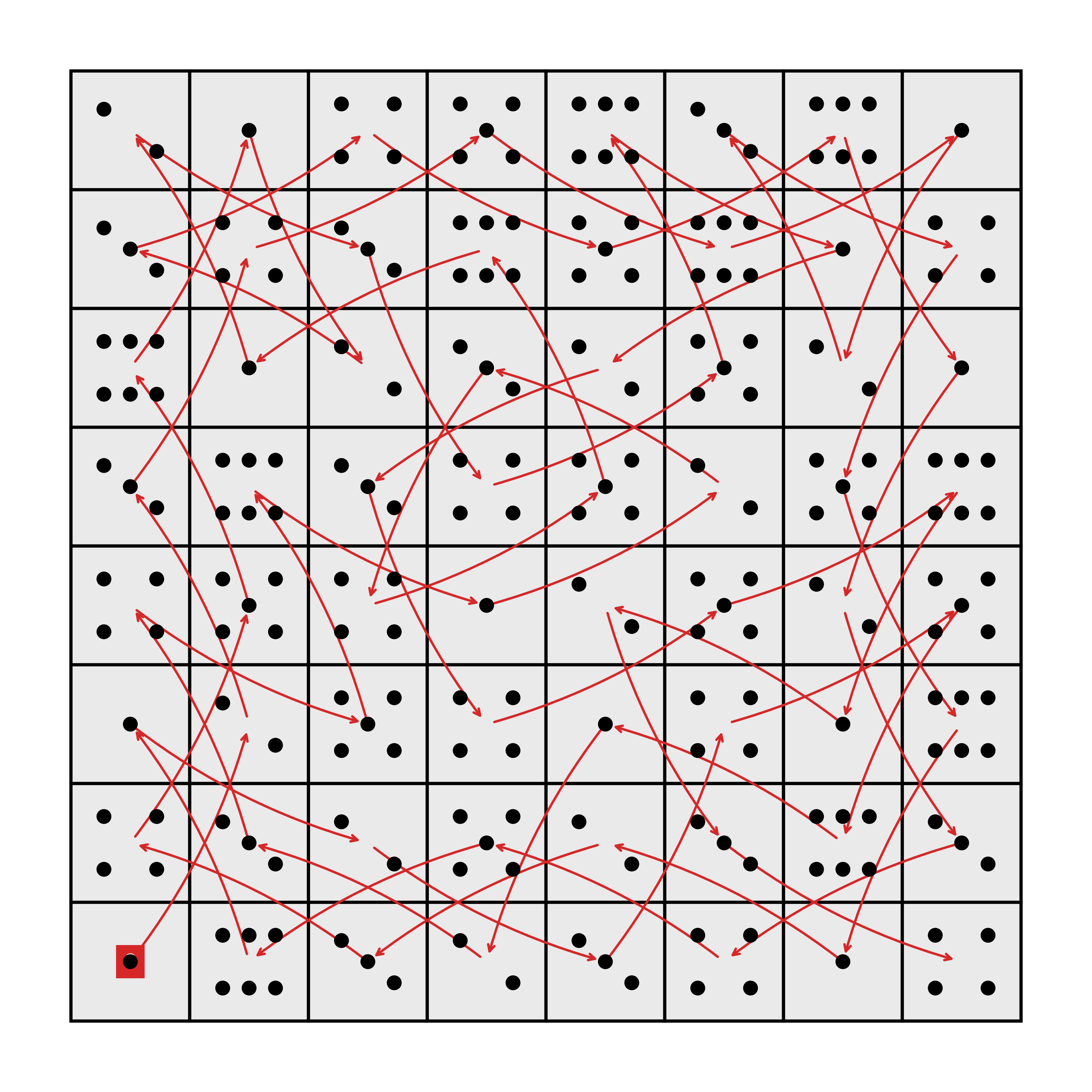}
    \captionof{figure}{A representative image, where poses a significant challenge to the perceptual abilities of current MLLMs.}
    \label{fig:hexagon_path}
  \end{center}
  \small
  \texttt{Model Response:}\textit{ Of course. Here is a detailed description of the image, designed to empower AI models without visual capabilities to reason and provide human experts with blueprint-level reconstructions.}
  \begin{enumerate}[label=\textbf{\arabic*.}, itemsep=8pt, leftmargin=*]
    \item \textit{
      \textbf{Overall Layout and Style}
      \newline
      This image is a square, minimalist abstract schematic diagram. Its core is a grid divided by black lines, with the grid cells filled in a light gray. Numerous solid black dots are scattered throughout the grid, a special starting point is marked by a red square, and there is a complex network of \change{numerous}{several} red arrows connecting different \change{black dots}{cells}. The overall image has no border, and the background is pure white.
    }
    \item \textit{
      \textbf{Coordinate System}
      \newline
      To precisely describe the position of all elements in the image, we will establish a Cartesian coordinate system based on the grid. The grid has an 8x8 structure. We will number the vertical columns from 1 to 8, moving from left to right, and the horizontal rows from 1 to 8, moving from bottom to top. Therefore, the cell in the bottom-left corner has the coordinates (1, 1), and the cell in the top-right corner has the coordinates (8, 8). When describing the black dots within each cell, we will use relative positional terms such as ``center," ``top-left," ``bottom-right," ``top-center," etc.
    }
    \item \textit{
      \textbf{Grid and Background}
      \newline
      The main body of the image is an 8x8 grid composed of thick black lines, containing a total of 64 perfectly identical square cells. The interior of each cell is filled with a light gray color (approximate hex code: \#E0E0E0).
    }
    \item \textit{
      \textbf{Detailed Element Description}
      \newline
      \begin{itemize}[itemsep=4pt, leftmargin=*]
      \item \textbf{4.1 Black Dots (Nodes)}
      \newline
      The image contains a total of \change{206}{216} uniformly sized, solid black dots (nodes). These nodes are distributed across the 64 grid cells, with the number and arrangement of nodes varying in each cell. One of these nodes is specially marked as the starting point. Below is a detailed distribution of the nodes in each cell, ordered from top to bottom (Row 8 to Row 1) and left to right (Column 1 to Column 8):
      \newline
      \textbf{Row 8 (Top)}
      \begin{itemize}
        \item \textbf{Cell (1, 8):} \change{1}{2} node, located at the \change{center.}{arranged diagonally, located at ``top-left" and ``bottom-right."}.
        \item \textbf{Cell (2, 8):} \change{3}{1} nodes, \change{arranged horizontally in the upper half of the cell. We will refer to them as ``left," ``middle," and ``right."}{located at the center.}
        \item \textbf{Cell (3, 8):} \change{3}{4} nodes, \change{arranged horizontally in the upper half of the cell. We will refer to them as ``left," ``middle," and ``right."}{arranged in a 2x2 square, located at ``top-left," ``top-right," ``bottom-left," and ``bottom-right."}
        \item \textbf{Cell (4, 8):} \change{3}{5} nodes, \change{arranged horizontally in the upper half of the cell. We will refer to them as ``left," ``middle," and ``right."}{arranged in a quincunx pattern, located at ``top-left," ``top-right," ``center," ``bottom-left," and ``bottom-right."}
        \item \textbf{Cell (5, 8):} \change{2}{6} nodes, \change{arranged diagonally, located at ``top-left" and ``bottom-right."}{arranged in two horizontal rows of three nodes each.}
        \item \textbf{Cell (6, 8):} 3 nodes, \change{arranged horizontally in the upper half of the cell. We will refer to them as ``left," ``middle," and ``right."}{arranged diagonally, located at ``top-left", ``center", and ``bottom-right."}
        \item \textbf{Cell (7, 8):} \change{3}{6} nodes, \change{arranged horizontally in the upper half of the cell. We will refer to them as ``left," ``middle," and ``right."}{arranged in two horizontal rows of three nodes each.}
        \item \textbf{Cell (8, 8):} \change{3}{1} nodes, \change{arranged horizontally in the upper half of the cell. We will refer to them as ``left," ``middle," and ``right."}{located at the center.}
      \end{itemize}        
      \textbf{Row 7}
      \begin{itemize}
        \item \textbf{Cell (1, 7):} \change{1}{3} node, \change{located at the center.}{arranged diagonally, located at ``top-left", ``center", and ``bottom-right."}
        \item \textbf{Cell (2, 7):} 4 nodes, arranged in a 2x2 square, located at ``top-left," ``top-right," ``bottom-left," and ``bottom-right."
        \item \textbf{Cell (3, 7):} 3 nodes, \change{forming a '<' shape pointing left, located at ``top-right," ``middle," and ``bottom-right."}{arranged diagonally, located at ``top-left", ``center", and ``bottom-right."}
        \item \textbf{Cell (4, 7):} \change{3}{6} nodes, \change{arranged in a pyramid shape, located at ``top-center," ``bottom-left," and ``bottom-right."}{arranged in two horizontal rows of three nodes each.}
        \item \textbf{Cell (5, 7):} \change{4}{5} nodes, \change{arranged in a 2x2 square, located at ``top-left," ``top-right," ``bottom-left," and ``bottom-right."}{arranged in a quincunx pattern, located at ``top-left," ``top-right," ``center," ``bottom-left," and ``bottom-right."}
        \item \textbf{Cell (6, 7):} \change{3}{6} nodes, \change{arranged in a pyramid shape, located at ``top-center," ``bottom-left," and ``bottom-right."}{arranged in two horizontal rows of three nodes each.}
        \item \textbf{Cell (7, 7):} \change{4}{1} nodes, \change{arranged in a 2x2 square, located at ``top-left," ``top-right," ``bottom-left," and ``bottom-right."}{located at the center.}
        \item \textbf{Cell (8, 7):} \change{3}{4} nodes, \change{forming a '<' shape pointing left, located at ``top-right," ``middle," and ``bottom-right."}{arranged in a 2x2 square, located at ``top-left," ``top-right," ``bottom-left," and ``bottom-right."}
      \end{itemize}
      \textbf{Row 6}
      \begin{itemize}
        \item \textbf{Cell (1, 6):} \change{1}{6} node, \change{located at the center.}{arranged in two horizontal rows of three nodes each.}
        \item \textbf{Cell (2, 6):} \change{3}{1} nodes, \change{arranged in a pyramid shape, located at ``top-center," ``bottom-left," and ``bottom-right."}{located at the center.}
        \item \textbf{Cell (3, 6):} \change{4}{2} nodes, \change{arranged in a 2x2 square, located at ``top-left," ``top-right," ``bottom-left," and ``bottom-right."}{arranged diagonally, located at ``top-left" and ``bottom-right."}
        \item \textbf{Cell (4, 6):} \change{4}{3} nodes, \change{arranged in a 2x2 square, located at ``top-left," ``top-right," ``bottom-left," and ``bottom-right."}{arranged diagonally, located at ``top-left", ``center", and ``bottom-right."}
        \item \textbf{Cell (5, 6):} \change{4}{2} nodes, \change{arranged in a 2x2 square, located at ``top-left," ``top-right," ``bottom-left," and ``bottom-right."}{arranged diagonally, located at ``top-left" and ``bottom-right."}
        \item \textbf{Cell (6, 6):} \change{3}{5} nodes, \change{arranged in a pyramid shape, located at ``top-center," ``bottom-left," and ``bottom-right."}{arranged in a quincunx pattern, located at ``top-left," ``top-right," ``center," ``bottom-left," and ``bottom-right."}
        \item \textbf{Cell (7, 6):} \change{3}{2} nodes, \change{arranged in a pyramid shape, located at ``top-center," ``bottom-left," and ``bottom-right."}{arranged diagonally, located at ``top-left" and ``bottom-right."}
        \item \textbf{Cell (8, 6):} \change{3}{1} nodes, \change{arranged in a pyramid shape, located at ``top-center," ``bottom-left," and ``bottom-right."}{located at the center.}
      \end{itemize}
      \textbf{Row 5}
      \begin{itemize}
        \item \textbf{Cell (1, 5):} \change{1}{3} node, \change{located at the center.}{arranged diagonally, located at ``top-left", ``center", and ``bottom-right."}
        \item \textbf{Cell (2, 5):} \change{4}{6} nodes, \change{arranged in a 2x2 square, located at ``top-left," ``top-right," ``bottom-left," and ``bottom-right."}{arranged in two horizontal rows of three nodes each.}
        \item \textbf{Cell (3, 5):} 3 nodes, \change{arranged in a pyramid shape, located at ``top-center," ``bottom-left," and ``bottom-right."}{arranged diagonally, located at ``top-left", ``center", and ``bottom-right."}
        \item \textbf{Cell (4, 5):} 4 nodes, arranged in a 2x2 square, located at ``top-left," ``top-right," ``bottom-left," and ``bottom-right."
        \item \textbf{Cell (5, 5):} \change{4}{5} nodes, \change{arranged in a 2x2 square, located at ``top-left," ``top-right," ``bottom-left," and ``bottom-right."}{arranged in a quincunx pattern, located at ``top-left," ``top-right," "center," ``bottom-left," and ``bottom-right."}
        \item \textbf{Cell (6, 5):} \change{3}{2} nodes, \change{arranged in a pyramid shape, located at ``top-center," ``bottom-left," and ``bottom-right."}{arranged diagonally, located at ``top-left" and ``bottom-right."}
        \item \textbf{Cell (7, 5):} \change{4}{5} nodes, \change{arranged in a 2x2 square, located at ``top-left," ``top-right," ``bottom-left," and ``bottom-right."}{arranged in a quincunx pattern, located at ``top-left," ``top-right," "center," ``bottom-left," and ``bottom-right."}
        \item \textbf{Cell (8, 5):} \change{3}{6} nodes, \change{arranged in a pyramid shape, located at ``top-center," ``bottom-left," and ``bottom-right."}{arranged in two horizontal rows of three nodes each.}
      \end{itemize}
      \textbf{Row 4}
      \begin{itemize}
        \item \textbf{Cell (1, 4):} \change{1}{4} node, \change{located at the center.}{arranged in a 2x2 square, located at ``top-left," ``top-right," ``bottom-left," and ``bottom-right."}
        \item \textbf{Cell (2, 4):} \change{3}{5} nodes, \change{arranged in a pyramid shape, located at ``top-center," ``bottom-left," and ``bottom-right."}{arranged in a quincunx pattern, located at ``top-left," ``top-right," "center," ``bottom-left," and ``bottom-right."}
        \item \textbf{Cell (3, 4):} 4 nodes, arranged in a 2x2 square, located at ``top-left," ``top-right," ``bottom-left," and ``bottom-right."
        \item \textbf{Cell (4, 4):} \change{4}{1} nodes, \change{arranged in a 2x2 square, located at ``top-left," ``top-right," ``bottom-left," and ``bottom-right."}{located at the center.}
        \item \textbf{Cell (5, 4):} \change{4}{2} nodes, \change{arranged in a 2x2 square, located at ``top-left," ``top-right," ``bottom-left," and ``bottom-right."}{arranged diagonally, located at ``top-left" and ``bottom-right."}
        \item \textbf{Cell (6, 4):} \change{4}{5} nodes, \change{arranged in a 2x2 square, located at ``top-left," ``top-right," ``bottom-left," and ``bottom-right."}{arranged in a quincunx pattern, located at ``top-left," ``top-right," "center," ``bottom-left," and ``bottom-right."}
        \item \textbf{Cell (7, 4):} \change{4}{2} nodes, \change{arranged in a 2x2 square, located at ``top-left," ``top-right," ``bottom-left," and ``bottom-right."}{arranged diagonally, located at ``top-left" and ``bottom-right."}
        \item \textbf{Cell (8, 4):} \change{4}{5} nodes, \change{arranged in a 2x2 square, located at ``top-left," ``top-right," ``bottom-left," and ``bottom-right."}{arranged in a quincunx pattern, located at ``top-left," ``top-right," "center," ``bottom-left," and ``bottom-right."}
      \end{itemize}
      \textbf{Row 3}
      \begin{itemize}
        \item \textbf{Cell (1, 3):} 1 node, located at the center.
        \item \textbf{Cell (2, 3):} \change{4}{2} nodes, \change{arranged in a 2x2 square, located at ``top-left," ``top-right," ``bottom-left," and ``bottom-right."}{arranged diagonally, located at ``top-left" and ``bottom-right."}
        \item \textbf{Cell (3, 3):} \change{3}{5} nodes, \change{arranged in a pyramid shape, located at ``top-center," ``bottom-left," and ``bottom-right."}{arranged in a quincunx pattern, located at ``top-left," ``top-right," "center," ``bottom-left," and ``bottom-right."}
        \item \textbf{Cell (4, 3):} 4 nodes, arranged in a 2x2 square, located at ``top-left," ``top-right," ``bottom-left," and ``bottom-right."
        \item \textbf{Cell (5, 3):} \change{4}{1} nodes, \change{arranged in a 2x2 square, located at ``top-left," ``top-right," ``bottom-left," and ``bottom-right."}{located at the center.}
        \item \textbf{Cell (6, 3):} \change{3}{4} nodes, \change{arranged in a pyramid shape, located at ``top-center," ``bottom-left," and ``bottom-right."}{arranged in a 2x2 square, located at ``top-left," ``top-right," ``bottom-left," and ``bottom-right."}
        \item \textbf{Cell (7, 3):} \change{4}{1} nodes, \change{arranged in a 2x2 square, located at ``top-left," ``top-right," ``bottom-left," and ``bottom-right."}{located at the center.}
        \item \textbf{Cell (8, 3):} \change{3}{6} nodes, \change{arranged in a pyramid shape, located at ``top-center," ``bottom-left," and ``bottom-right."}{arranged in two horizontal rows of three nodes each.}
      \end{itemize}
      \textbf{Row 2}
      \begin{itemize}
        \item \textbf{Cell (1, 2):} \change{1}{4} node, \change{located at the center.}{arranged in a 2x2 square, located at ``top-left," ``top-right," ``bottom-left," and ``bottom-right."} \change{\textbf{This is the special starting point.}}{From (1, 1) to (2, 3)}
        \item \textbf{Cell (2, 2):} 3 nodes, \change{arranged in a pyramid shape, located at ``top-center," ``bottom-left," and ``bottom-right."}{arranged diagonally, located at ``top-left", ``center", and ``bottom-right."}
        \item \textbf{Cell (3, 2):} \change{4}{2} nodes, \change{arranged in a 2x2 square, located at ``top-left," ``top-right," ``bottom-left," and ``bottom-right."}{arranged diagonally, located at ``top-left" and ``bottom-right."}
        \item \textbf{Cell (4, 2):} \change{4}{5} nodes, \change{arranged in a 2x2 square, located at ``top-left," ``top-right," ``bottom-left," and ``bottom-right."}{arranged in a quincunx pattern, located at ``top-left," ``top-right," "center," ``bottom-left," and ``bottom-right."}
        \item \textbf{Cell (5, 2):} \change{4}{2} nodes, \change{arranged in a 2x2 square, located at ``top-left," ``top-right," ``bottom-left," and ``bottom-right."}{arranged diagonally, located at ``top-left" and ``bottom-right."}
        \item \textbf{Cell (6, 2):} \change{4}{3} nodes, \change{arranged in a 2x2 square, located at ``top-left," ``top-right," ``bottom-left," and ``bottom-right."}{arranged diagonally, located at ``top-left", ``center", and ``bottom-right."}
        \item \textbf{Cell (7, 2):} \change{4}{6} nodes, \change{arranged in a 2x2 square, located at ``top-left," ``top-right," ``bottom-left," and ``bottom-right."}{arranged in two horizontal rows of three nodes each.}
        \item \textbf{Cell (8, 2):} \change{4}{3} nodes, \change{arranged in a 2x2 square, located at ``top-left," ``top-right," ``bottom-left," and ``bottom-right."}{arranged diagonally, located at ``top-left", ``center", and ``bottom-right."}
      \end{itemize}
      \textbf{Row 1 (Bottom)}
      \begin{itemize}
        \item \textbf{Cell (1, 1):} 1 node, located at the center. \textbf{This is the special starting point.}
        \item \textbf{Cell (2, 1):} \change{3}{6} nodes, \change{arranged horizontally in the lower half of the cell. We will refer to them as ``left," ``middle," and ``right."}{arranged in two horizontal rows of three nodes each.}
        \item \textbf{Cell (3, 1):} \change{4}{3} nodes, \change{arranged in a 2x2 square, located at ``top-left," ``top-right," ``bottom-left," and ``bottom-right."}{arranged diagonally, located at ``top-left", ``center", and ``bottom-right."}
        \item \textbf{Cell (4, 1):} \change{3}{2} nodes, \change{arranged horizontally in the lower half of the cell. We will refer to them as ``left," ``middle," and ``right."}{arranged diagonally, located at ``top-left" and ``bottom-right."}
        \item \textbf{Cell (5, 1):} \change{4}{3} nodes, \change{arranged in a 2x2 square, located at ``top-left," ``top-right," ``bottom-left," and ``bottom-right."}{arranged diagonally, located at ``top-left", ``center", and ``bottom-right."}
        \item \textbf{Cell (6, 1):} \change{3}{4} nodes, \change{arranged horizontally in the lower half of the cell. We will refer to them as ``left," ``middle," and ``right."}{arranged in a 2x2 square, located at ``top-left," ``top-right," ``bottom-left," and ``bottom-right."}
        \item \textbf{Cell (7, 1):} \change{4}{1} nodes, \change{arranged in a 2x2 square, located at ``top-left," ``top-right," ``bottom-left," and ``bottom-right."}{located at the center.}
        \item \textbf{Cell (8, 1):} 4 nodes, arranged in a 2x2 square, located at ``top-left," ``top-right," ``bottom-left," and ``bottom-right."
      \end{itemize}
      \item \textbf{4.2 Starting Point}
      \newline
      In cell (1, 1), the single central node is specially marked as the starting point. It is enclosed by a square slightly smaller than the cell itself. This square is filled with a bright red color (approximate hex code: \#D82428) \change{and has a thick black border.}{.}
      \item \textbf{4.3 Red Arrows (Connections)}
      \newline
      There are a total of \change{\textbf{68}}{\textbf{63}} red arrows in the image. Each arrow is a straight line originating from the center of one black node and pointing to the center of another, with an arrowhead at its tip to indicate direction. These arrows crisscross to form a complex network. The following is a complete list of all \change{68}{63} arrows, describing their origin and destination points:
      \begin{enumerate}[label=\arabic*., itemsep=2pt, leftmargin=*]
        \item \change{From (1, 1, center) to (2, 3, top-left)}{From (1, 1) to (2, 3)}
        \item \change{From (1, 2, center) (Start Point) to (2, 1, left)}{From (2, 3) to (1, 5)}
        \item \change{From (1, 5, center) to (2, 4, top-center)}{From (1, 5) to (2, 7)}
        \item \change{From (1, 6, center) to (2, 8, left)}{From (2, 7) to (4, 8)}
        \item \change{From (1, 7, center) to (2, 7, bottom-right)}{From (4, 8) to (6, 7)}
        \item \change{From (1, 8, center) to (1, 7, center)}{From (6, 7) to (8, 8)}
        \item \change{From (2, 1, left) to (1, 1, center)}{From (8, 8) to (7, 6)}
        \item \change{From (2, 2, bottom-left) to (3, 3, top-center)}{From (7, 6) to (6, 8)}
        \item \change{From (2, 2, bottom-right) to (1, 3, center)}{From (6, 8) to (8, 7)}
        \item \change{From (2, 3, top-left) to (3, 4, bottom-left)}{From (8, 7) to (7, 5)}
        \item \change{From (2, 4, top-center) to (2, 6, top-center)}{From (7, 5) to (8, 3)}
        \item \change{From (2, 4, bottom-right) to (3, 2, top-right)}{From (8, 3) to (7, 1)}
        \item \change{From (2, 5, bottom-right) to (1, 6, center)}{From (7, 1) to (5, 2)}
        \item \change{From (2, 6, top-center) to (1, 5, center)}{From (5, 2) to (3, 1)}
        \item \change{From (2, 7, bottom-right) to (3, 7, middle)}{From (3, 1) to (1, 2)}
        \item \change{From (2, 8, left) to (1, 8, center)}{From (1, 2) to (2, 4)}
        \item \change{From (3, 1, top-right) to (4, 3, top-left)}{From (2, 4) to (1, 6)}
        \item \change{From (3, 2, top-right) to (4, 1, right)}{From (1, 6) to (2, 8)}
        \item \change{From (3, 2, bottom-right) to (2, 2, bottom-left)}{From (2, 8) to (3, 6)}
        \item \change{From (3, 3, top-center) to (4, 3, bottom-right)}{From (3, 6) to (1, 7)}
        \item \change{From (3, 4, bottom-left) to (4, 5, bottom-right)}{From (1, 7) to (3, 8)}
        \item \change{From (3, 5, bottom-right) to (4, 7, top-left)}{From (3, 8) to (5, 7)}
        \item \change{From (3, 6, left) to (2, 5, bottom-right)}{From (5, 7) to (7, 8)}
        \item \change{From (3, 7, middle) to (4, 8, left)}{From (7, 8) to (8, 6)}
        \item \change{From (3, 7, top-right) to (2, 6, bottom-right)}{From (8, 6) to (7, 4)}
        \item \change{From (4, 1, left) to (3, 1, top-right)}{From (7, 4) to (8, 2)}
        \item \change{From (4, 2, top-left) to (3, 1, bottom-left)}{From (8, 2) to (6, 1)}
        \item \change{From (4, 3, top-left) to (5, 5, top-left)}{From (6, 1) to (4, 2)}
        \item \change{From (4, 3, bottom-right) to (5, 4, bottom-right)}{From (4, 2) to (2, 1)}
        \item \change{From (4, 4, top-left) to (5, 3, top-left)}{From (2, 1) to (1, 3)}
        \item \change{From (4, 5, bottom-right) to (3, 6, left)}{From (1, 3) to (3, 2)}
        \item \change{From (4, 5, top-right) to (3, 5, top-center)}{From (3, 2) to (5, 1)}
        \item \change{From (4, 6, bottom-left) to (3, 5, bottom-right)}{From (5, 1) to (6, 3)}
        \item \change{From (4, 6, top-right) to (5, 8, bottom-right)}{From (6, 3) to (8, 4)}
        \item \change{From (4, 7, top-left) to (5, 7, bottom-left)}{From (8, 4) to (7, 2)}
        \item \change{From (4, 8, left) to (5, 8, top-left)}{From (7, 2) to (5, 3)}
        \item \change{From (5, 1, middle-right) to (4, 2, top-left)}{From (5, 3) to (4, 1)}
        \item \change{From (5, 2, bottom-left) to (6, 1, right)}{From (4, 1) to (2, 2)}
        \item \change{From (5, 3, top-left) to (6, 3, bottom-right)}{From (2, 2) to (1, 4)}
        \item \change{From (5, 4, bottom-right) to (4, 4, top-left)}{From (1, 4) to (3, 3)}
        \item \change{From (5, 4, top-left) to (4, 4, top-right)}{From (3, 3) to (2, 5)}
        \item \change{From (5, 5, bottom-right) to (5, 2, bottom-left)}{From (2, 5) to (4, 4)}
        \item \change{From (5, 5, top-left) to (4, 6, bottom-left)}{From (4, 4) to (6, 5)}
        \item \change{From (5, 6, top-right) to (4, 6, top-right)}{From (6, 5) to (4, 6)}
        \item \change{From (5, 7, bottom-left) to (6, 7, top-center)}{From (4, 6) to (3, 4)}
        \item \change{From (5, 8, top-left) to (6, 8, left)}{From (3, 4) to (5, 5)}
        \item \change{From (6, 1, middle) to (5, 1, middle-right)}{From (5, 5) to (4, 7)}
        \item \change{From (6, 2, bottom-left) to (5, 1, top-left)}{From (4, 7) to (2, 6)}
        \item \change{From (6, 3, bottom-right) to (7, 3, top-right)}{From (2, 6) to (1, 8)}
        \item \change{From (6, 4, top-right) to (7, 4, bottom-right)}{From (1, 8) to (3, 7)}
        \item \change{From (6, 5, bottom-left) to (5, 6, top-right)}{From (3, 7) to (4, 5)}
        \item \change{From (6, 6, top-right) to (5, 5, bottom-right)}{From (4, 5) to (6, 6)}
        \item \change{From (6, 7, top-center) to (7, 7, bottom-right)}{From (6, 6) to (5, 8)}
        \item \change{From (6, 8, left) to (7, 8, left)}{From (5, 8) to (7, 7)}
        \item \change{From (7, 1, top-left) to (6, 2, bottom-left)}{From (7, 7) to (5, 6)}
        \item \change{From (7, 2, bottom-right) to (8, 2, bottom-left)}{From (5, 6) to (3, 5)}
        \item \change{From (7, 3, top-right) to (6, 5, bottom-left)}{From (3, 5) to (4, 3)}
        \item \change{From (7, 4, bottom-right) to (8, 3, top-right)}{From (4, 3) to (6, 4)}
        \item \change{From (7, 5, bottom-left) to (6, 4, top-right)}{From (6, 4) to (8, 5)}
        \item \change{From (7, 6, bottom-left) to (8, 6, bottom-right)}{From (8, 5) to (7, 3)}
        \item \change{From (7, 7, bottom-right) to (6, 6, top-right)}{From (7, 3) to (5, 4)}
        \item \change{From (7, 8, left) to (8, 8, left)}{From (5, 4) to (6, 2)}
        \item \change{From (8, 1, top-right) to (7, 1, top-left)}{From (6, 2) to (8, 1)}
        \item \change{From (8, 2, bottom-left) to (8, 1, top-right)}{}
        \item \change{From (8, 3, top-right) to (7, 2, bottom-right)}{}
        \item \change{From (8, 5, bottom-left) to (7, 5, bottom-left)}{}
        \item \change{From (8, 6, bottom-right) to (8, 5, bottom-left)}{}
        \item \change{From (8, 7, bottom-right) to (7, 6, bottom-left)}{}
      \end{enumerate}
    \end{itemize}
    }
    \textit{The above description provides the precise number, location, shape, attributes, and interrelationships of all visual elements in the image, which is sufficient to support in-depth analysis or accurate reproduction of the image.}
  \end{enumerate}
\end{takeawaybox}

\section{Training Details}
\label{supp:Training_Details}
\subsection*{1 The Training Details of CodePercept-S1}
CodePercept-S1 is built upon the Qwen3-VL-Instruct foundation model and fine-tuned using a carefully curated mixture of dataset ICC-1M, which is derived from public STEM seed data including Mulberry-SFT~\cite{yao2024mulberry} and GRAMMAR~\cite{chen2025advancing}. The model is trained for 1 epochs using AdamW optimizer with DeepSpeed Zero-2 optimization, employing a cosine learning rate schedule starting from $3\times10^{-6}$ and decaying to $2\times10^{-7}$ with 5\% warm-up steps. Training is conducted with an effective batch size of 4 per device (2 samples per device with gradient accumulation of 2 steps), using bfloat16 mixed precision and Flash Attention for computational efficiency. 
\subsection*{2 The Training Details of CodePercept-R1}
CodePercept-R1 advances upon CodePercept-S1 through Group Relative Policy Optimization (GRPO), a reinforcement learning approach designed to enhance code generation quality. The model is initialized from the CodePercept-S1 checkpoint and further refined using a carefully selected subset of 10,000 high-quality samples from the ICC-1M dataset, chosen based on quality filter. 
We first perform a one-time difficulty filtering based on Qwen3-VL-7B: under 8 rollout iterations, we select samples whose accuracy falls between 0.25 and 0.75~\cite{yu2025dapo}. We utilize GRPO with a temperature of 1.0, top-p sampling of 0.85, and a repetition penalty of 1.1 to encourage diverse and coherent generations. For each training sample, the model generates 8 candidate completions with a maximum length of 8,192 tokens, which are then optimized using a composite reward function.
Training is conducted for 1 epoch using the AdamW optimizer with a learning rate of $1\times10^{-6}$ and 5\% warm-up steps. The optimization uses bfloat16 mixed precision and gradient clipping with a maximum norm of 0.5 for training stability. The GRPO algorithm uses a KL penalty coefficient (beta) of 0.001 to balance between reward optimization and maintaining proximity to the reference policy.
\subsection*{3 The Training Curves of CodePercept-S1 series and CodePercept-R1 series}
\begin{figure}[t]
    \centering
    \includegraphics[width=1.0\linewidth]{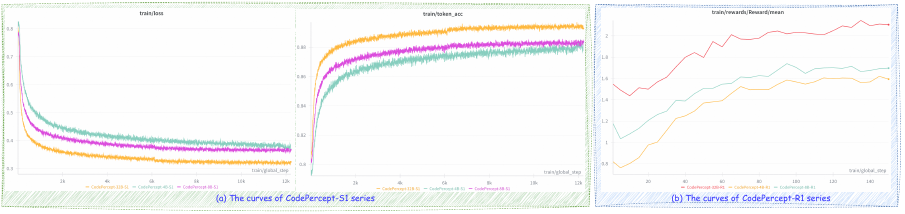}
    \caption{The training curves of our proposed models. (a) the curves of CodePercept-S1 models. (b) the curves of CodePercept-R1 models.}
    \label{supp:training_details}
\end{figure}

As shown in Fig.~\ref{supp:training_details}, this image displays three charts illustrating the training performance of different versions of a model named ``CodePercept". The charts are divided into two groups: (a) for the ``CodePercept-S1 series" and (b) for the ``CodePercept-R1 series". Both series compare models of three different sizes: 4 billion (4B), 8 billion (8B), and 32 billion (32B) parameters. Across both the S1 and R1 training stages, the charts consistently demonstrate a clear scaling law: larger models in the CodePercept family exhibit better performance. Whether in the supervised fine-tuning stage (S1) or the reinforcement learning stage (R1), the CodePercept model with a larger number of parameters (32B $>$ 8B $>$ 4B) demonstrates superior performance, namely lower loss, higher accuracy, and higher reward value.

\section{The Construction Pipeline of STEM2Code-Eval Benchmark}
\label{supp:STEM2Code-Eval}
\begin{figure}[t]
    \centering
    \includegraphics[width=\linewidth]{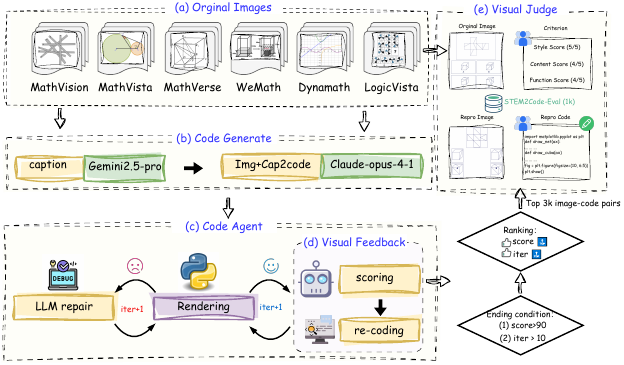}
    \caption{The pipeline of our proposed STEM2Code-Eval Benchmark, which includes 1000 image-code pairs that have been checked and revised by human annotators.}
    \label{fig:STEM2Code}
\end{figure}
In the section, we provide the specific construction pipeline of our STEM2Code-Eval benchmark, as shown in Fig~\ref{supp:STEM2Code-Eval}. The caption prompt is the same as \texttt{Native Caption Prompt} (see Sec.\ref{supp:Code_Grounded_Caption}),
and the img+cap2code prompt is foremulated as follows:
\begin{takeawaybox}{Img+Cap2Code prompt}
  \small
  \textit{You are a top-tier programmer proficient in Python and Matplotlib.}
  \begin{enumerate}[label=\textbf{\arabic*.}, itemsep=8pt, leftmargin=*]
    \item \textit{
      \textbf{Primary Objective}
      \newline
      \textbf{Task}: Based on an input image and its corresponding detailed image description, generate a standalone, high-quality Python code to reproduce that image.
      \newline
      The precise description of the input image is:
      \newline
      \texttt{[description]}
      \begin{enumerate}[label=\alph*., itemsep=4pt, leftmargin=*]
        \item \textbf{Core Requirement}: Observe the image, and based on the \textbf{image description} you received and the extracted \textbf{core principles}, precisely reproduce the original image.
        \item \textbf{Visual Matching}: Must ensure complete consistency with the following aspects in the description:
          \begin{itemize}
            \item Coordinate system range and scales
            \item Colors, line widths, and styles
            \item Element positions and sizes
            \item If the description includes axes, ticks, grids, etc., they must be precisely reproduced.
          \end{itemize}
        \item \textbf{Code Algorithm}: The code should not only generate the correct image, but also be written in a clear, structured manner that is intuitively reflected in the final image. The purpose is: An observer should be able to analyze the parameters or logic in the code by merely observing the rendered image, ultimately achieving Image2Code restoration.
      \end{enumerate}
    }
    \item \textit{
      \textbf{Output Format \& Prohibitions}
      \begin{itemize}
        \item Your entire response must be a \textbf{single text block} with the following structure: \newline \textbf{Python Code Block}: Strictly enclosed by \texttt{```python ```}.
        \item Code is self-contained with import, one main function, and one main entry point.
        \item Code must include \texttt{plt.show()} at the end.
      \end{itemize}
    }
  \end{enumerate}
\end{takeawaybox}
Next, the LLM repair prompt is define as:
\begin{takeawaybox}{LLM Repair Prompt}
  \small
  \textit{You are a Code Debug Assistant. Your task is to identify and fix issues in the user's code based on any provided errors, ensuring it works correctly. You will be given the user's code and a corresponding error message in the following format:\newline
      \#\#\# Error code\newline
    \texttt{```python [Error code] ```}\newline
    \#\#\# Error message\newline
    \texttt{```text [Error message] ```}\newline
        Your response must strictly adhere to the following criteria:
          \begin{itemize}
            \item Return ONLY the complete, corrected Python code.
            \item The code must be enclosed within a single \texttt{```python...```} block.
            \item You must return all the complete, working code, not just the modified part.
          \end{itemize}
}
\end{takeawaybox}
Next, the scoring prompt is defined as follows:
\label{supp:Image Scoring Prompt}
\begin{takeawaybox}{Image Scoring Prompt}
  \small
  \textit{You are an expert judge in evaluating mathematical and geometric diagrams. The first image (reference image) is a ground truth mathematical figure, and the second image (AI-generated image) is created using code generated by an AI assistant. Your task is to score how well the AI-generated image matches the ground truth image.}
  \begin{enumerate}[label=\textbf{\arabic*.}, itemsep=8pt, leftmargin=*]
    \item \textit{
      \textbf{Scoring Methodology}
      \newline
      \textbf{Task}: The AI-generated image's score is based on the following criteria, totaling a score out of 100 points. The evaluation must consider the mathematical and geometric correctness of the figure, focusing on the precise arrangement and relationships of its components.
      \newline
      \begin{enumerate}[label=\textbf{Criterion \arabic*.}, itemsep=4pt, leftmargin=*]
        \item \textbf{Geometric \& Structural Completeness (30 points)}:
          \begin{itemize}
            \item \textbf{Element Types:} Does the AI-generated image include all fundamental element types from the reference image (e.g., points, lines, segments, rays, circles, polygons, curves, coordinate axes, text labels)?
            \item \textbf{Element Quantity:} Is the \textbf{exact number} of each element type correct? (e.g., if the reference has 8 points and 3 triangles, does the generated image also have exactly 8 points and 3 triangles?).
          \end{itemize}
        \item \textbf{Positional \& Relational Accuracy (30 points)}:
          \begin{itemize}
            \item \textbf{Absolute \& Relative Positioning:} Are all elements placed at their correct locations? This assesses accuracy within the image's implicit or explicit coordinate system.
            \item \textbf{Spatial Relationships:} Does the image correctly represent all spatial relationships, such as \textbf{adjacency}, \textbf{intersection}, \textbf{containment}, \textbf{collinearity}, and \textbf{parallelism/perpendicularity}?
            \item \textbf{Sequential \& Topological Relationships:} For figures like graphs or paths, is the \textbf{sequence of connections} correct? Is the overall topological structure preserved?
            \item \textbf{Layering (Z-order):} Are overlapping elements stacked in the correct order (e.g., is the shaded region correctly drawn behind the boundary line)?
          \end{itemize}
        \item \textbf{Text \& Annotation Fidelity (10 points)}:
            \begin{itemize}
                \item \textbf{Content:} Does the AI-generated image include all text and symbolic annotations from the reference? Is the content of the text identical?
                \item \textbf{Positioning \& Association:} Are annotations placed correctly relative to the geometric elements they describe?
                \item \textbf{Style:} Does the style of the text (e.g., font, size, italics for variables, use of mathematical symbols) match the reference?
            \end{itemize}
        \item \textbf{Visual \& Stylistic Consistency (20 points)}:
            \begin{itemize}
                \item \textbf{Colors \& Fill:} Do the colors (stroke, fill) of all elements match the reference? Are shaded regions filled correctly?
                \item \textbf{Line \& Marker Styles:} Do line styles (e.g., solid, dashed, dotted), line weights, and marker styles (e.g., dots, small circles, crosses) match?
                \item \textbf{Overall Aesthetics:} Does the overall appearance, including background color, grid lines, and aspect ratio, match the reference image?
            \end{itemize}
        \item \textbf{Clarity \& Legibility (10 points)}:
            \begin{itemize}
                \item Is the AI-generated image clear, sharp, and well-rendered?
                \item Are there any distracting artifacts, incorrect overlaps, or elements that are difficult to distinguish? Is all text legible?
            \end{itemize}
      \end{enumerate}
    }
    \item \textit{
      \textbf{Evaluation \& Response Format}
      \begin{itemize}[itemsep=4pt]
        \item Compare the two images head to head and provide a detailed assessment based on the criteria above.
        \item Use the following format for your response to ensure the evaluation is clear and comprehensive.
      \end{itemize}
      \begin{itemize}[leftmargin=2em, itemsep=1pt, label={}]
          \item \texttt{---}
          \item \texttt{Comments:}
          \item \texttt{- Geometric \& Structural Completeness: \$\{your comment and subscore\}}
          \item \texttt{- Positional \& Relational Accuracy: \$\{your comment and subscore\}}
          \item \texttt{- Text \& Annotation Fidelity: \$\{your comment and subscore\}}
          \item \texttt{- Visual \& Stylistic Consistency: \$\{your comment and subscore\}}
          \item \texttt{- Clarity \& Legibility: \$\{your comment and subscore\}}
          \item
          \item \texttt{Score: \$\{your final score out of 100\}}
          \item \texttt{---}
      \end{itemize}
      Please use the above format to ensure the evaluation is clear and comprehensive.
    }
  \end{enumerate}
\end{takeawaybox}
Next, the re-coding prompt is defined as:
\begin{takeawaybox}{Re-Scoring Prompt}
  \small
  \textit{You are an expert in Python data visualization, skilled at accurately recreating scientific charts through visual comparison.}\newline
  \textit{
You will receive three inputs:\newline
\texttt{[Original Image]}: The first image, our target for reproduction, originating from the fields of Science, Technology, Engineering, and Mathematics (STEM). \newline
\texttt{[Current Render]}: The second image, a preliminary version generated by the code below. \newline
\texttt{[Current Code]}: The Python code that generated the ``Current Render". \newline
\texttt{```python [Current Code] ```}.
\newline
Your core task is to identify the differences between the ``Current Render" and the ``Original Image" in terms of core scientific information and key structures, and then help me correct the code to reproduce the original image as closely as possible.
    }
\end{takeawaybox}
\section{The Evaluation Implementation of STEM2Code-Eval Benchmark}
In this section, we provide a detailed description of the implementation for the three metrics used in our STEM2Code-Eval benchmark: Image Score, Code Score, and Execution Rate (Exec Rate).
\subsection{Image Score}
The Image Score is designed to evaluate the visual fidelity and semantic correctness of the generated image compared to the ground-truth image. We leverage the advanced multi-modal understanding capabilities of Google's Gemini2.5-Pro for this task.
The evaluation process is as follows:
\begin{enumerate}
    \item \textbf{Input}: For each sample, we provide Gemini2.5 Pro with both the ground-truth image and the image generated by the model-under-test's code.
    \item \textbf{Prompting}: We use a carefully crafted prompt that instructs the model to act as an expert evaluator. The prompt asks for a comprehensive comparison focusing on key visual elements, data representation accuracy, color schemes, layout, and overall similarity.
    \item \textbf{Scoring}: The model is instructed to provide a final score on a scale of 0 to 100, where 0 indicates no resemblance and 100 indicates a pixel-perfect match. To ensure reliable parsing, we require the score to be formatted in a specific way (e.g., ``Final Score: [score]").
    \item \textbf{Extraction}: We programmatically parse the model's textual response to extract the numerical score for aggregation.
\end{enumerate}

We used the same prompt for Gemini 2.5 Pro as described in Sec.\ref{supp:Image Scoring Prompt}.
\subsection{Code Score}
The Code Score assesses the quality of the generated Python code itself. This evaluation goes beyond mere executability and focuses on correctness, readability, and best practices. We utilize OpenAI's GPT-4o for its strong code understanding and generation capabilities.
The evaluation process is as follows:
\begin{enumerate}
    \item \textbf{Input}: For each sample, we provide GPT-4o with the Python code generated by the model-under-test.
    \item \textbf{Prompting}: The model is prompted to act as a senior software engineer conducting a code review. The prompt specifies multiple dimensions for evaluation: logical correctness, readability, efficiency, and robustness (e.g., handling of imports).
    \item \textbf{Scoring}: The model is asked to provide a score for each dimension and then an overall quality score on a scale of 0 to 100. This multi-faceted approach provides a more nuanced assessment.
    \item \textbf{Extraction}: The final overall score is extracted from the formatted response.
\end{enumerate}
The specific prompt used for GPT-4o is as follows:
\begin{takeawaybox}{Code Scoring Prompt}
\textit{You will act as an expert judge, responsible for rigorous visual verification of AI-generated graphics code.
Your sole task is to evaluate whether the AI code is completely consistent with the reference code written by human experts in terms of the final rendered visual result. You must ignore technical differences in the code implementation (e.g., algorithms, data structures) and focus on every pixel and geometric detail that goes into rendering the final image.}
\newline
\begin{enumerate}[label=\textbf{\arabic*.}, itemsep=4pt, leftmargin=*]
\item \textit{
\textbf{Evaluation Mission and Core Principles}
    \begin{enumerate}[label=\textbf{Criterion \arabic*.}, itemsep=4pt, leftmargin=*]
    \item Visual Identity: Two pieces of code that render the exact same image should be considered equally valid. Elegance or clumsiness of the implementation is irrelevant to the scoring.
    \item Pixel-Level Accuracy: Your evaluation must be accurate down to the pixel level. This includes geometric shape outlines, position, number of elements, relative relationships, and all visual attributes.
    \item Objective and Quantitative: All comments must be supported by concrete visual evidence, strictly adhering to the following scoring criteria.
    \item Unconditional Evaluation: The evaluation must be performed on any provided AI-generated code, \textbf{regardless of whether it is empty or incomplete}. You must be scored accordingly by applying the standard criteria, which will naturally result in a very low score.
    \end{enumerate}
}
\item \textit{
\textbf{Scoring Criteria and Guidelines (100 points)}\newline
You will be scored based on the following five criteria. Each item is directly related to the final visual presentation.
\begin{enumerate}[label=\textbf{Criterion \arabic*.}, itemsep=4pt, leftmargin=*]
\item Overall Layout and Visual Attribute Fidelity (20 points)
\begin{itemize}
    \item Canvas and Coordinate System: Are the canvas attributes (e.g., aspect ratio, background color) correct? If a grid or coordinate system exists, are its range, scale, and scale consistent with the reference standard?
    \item Macro Layout: Is the overall basic framework of the graphic correct? (For example, where is the main subject located on the canvas, and is the overall visual center of gravity consistent?)
    \item Color and Style: Are the fill color, stroke color, and opacity of all elements consistent with the reference code? Do the line width, style (solid, dashed, dotted), and cap style (round, square) match?
    \item Text and Annotations: If text labels or mathematical annotations exist, are their content, font, size, position, and alignment consistent with the reference code?
\end{itemize}
\item Quantitative Fidelity (20 points)
    \begin{itemize}
    \item Element List Verification: Does the AI-generated graphic contain the exact same types and numbers of geometric elements as the reference code? (e.g., 8 polygons, 14 path nodes, 1 mesh).
    \item Completeness: Are there any missing or redundant geometric components compared to the reference code?
    \end{itemize}
\item Positioning and Layout Accuracy (30 points)
    \begin{itemize}
    \item Absolute Coordinate Accuracy: Do the coordinates of all key elements (e.g., polygon vertices, circle centers, path anchor points) precisely match those calculated in the reference code?
    \item Relative Position Relationship: Are the spatial arrangement of elements correct? (e.g., A is above and left of B, C and D are horizontally aligned, and a group of elements are arranged in a circular pattern).
    \item Alignment and Distribution: Do the elements follow the same alignment (left/right/center) and distribution (uniform/non-uniform) pattern as the reference code?
    \end{itemize}
\item Relationship and Stacking Completeness (20 points)
    \begin{itemize}
    \item Connectivity and Sequence: If the graph contains paths, networks, or ordered sequences, is the order of connections between nodes **perfectly reproduced**? Are the starting and ending points of lines or paths correct?
    \item Spatial Interaction: Are complex relationships between elements (such as adjacency, containment, intersection, and overlap) rendered correctly? Are the shapes and sizes of overlapping areas accurate?
    \item Stacking Order (Z-index): When elements overlap, are they stacked in the correct order (i.e., which element is on top and which is on the bottom)?
    \end{itemize}
\item Code Implementation and Quality (10 points)
    \begin{itemize}
    \item Clarity and Readability: Is the code well-structured and clear? Does it use meaningful variable names and appropriate comments?
    \item Correctness and Efficiency: Is the code free of syntactical errors, logical errors, and unnecessary redundancy? Does it effectively use appropriate functions and methods from relevant libraries?
    \item Reproducibility: When executed in the correct environment, does the code run correctly and produce the expected complete graph? ---
    \end{itemize}
\end{enumerate}
}
\item \textit{\textbf{Evaluation}\newline
Compare the reference code to the AI code.
Provide a detailed evaluation and rating for each criterion, and then calculate a final overall score.\newline
You must strictly adhere to the following format for your response. (Highest Priority)
}
\begin{itemize}[leftmargin=2em, itemsep=1pt, label={}]
  \item \texttt{---}
  \item \texttt{Comments:}
  \item \texttt{- Geometric \& Structural Completeness: \$\{your comment and subscore\}}
  \item \texttt{- Positional \& Relational Accuracy: \$\{your comment and subscore\}}
  \item \texttt{- Text \& Annotation Fidelity: \$\{your comment and subscore\}}
  \item \texttt{- Visual \& Stylistic Consistency: \$\{your comment and subscore\}}
  \item \texttt{- Clarity \& Legibility: \$\{your comment and subscore\}}
  \item
  \item \texttt{Score: \$\{your final score out of 100\}}
  \item \texttt{---}
\end{itemize}
\end{enumerate}
\textit{Now, give your reference code and AI-generated code in the following format:\newline
\#\#\# Reference Code\newline
\texttt{```python [Reference Code] ```}\newline
\#\#\# AI-Generated Code\newline
\texttt{```python [AI-Generated Code] ```}\newline
Please use the above format to ensure the evaluation is clear and comprehensive.
}
\end{takeawaybox}

\subsection{Exec Rate}
The Execution Rate (Exec Rate) is a binary metric that measures the direct success rate of the generated code. It is the most objective of the three metrics. A code snippet is considered successful if and only if it executes without errors and produces the intended visual output file.

The automated execution pipeline is implemented as follows:
\begin{enumerate}
\item \textbf{Sandboxed Environment}: Each code snippet is executed in an isolated and clean sandboxed environment. This ensures that executions do not interfere with each other and provides a consistent environment. The environment is pre-configured with a standard set of Python data science libraries, including numpy, matplotlib, scipy, pandas, etc. To prevent rendering failures, necessary font libraries for handling Chinese characters and icons are also pre-installed.
\item \textbf{Execution}: The generated Python code is saved to a file (e.g., run.py) and executed using a Python interpreter. A timeout limit of 120 seconds is imposed to handle cases of infinite loops or excessively long computations.
\item \textbf{Verification}: Success is determined by a two-fold check:
\begin{itemize}
\item The script must complete with an exit code of 0, indicating no runtime errors.
\item The script must generate an image file (e.g., .png, .jpg) in the working directory. The presence of this artifact is checked post-execution.
\end{itemize}
\item \textbf{Calculation}: The Exec Rate is calculated as the percentage of samples that pass both verification checks:
$\text{Exec Rate} = \frac{\text{Number of Successfully Executed Samples}}{\text{Total Number of Samples}} \times 100$
\end{enumerate}

\section{Experiments with 32B}
In this section, we continue to validate the effectiveness of our proposed CodePercept method in public six STEM benchmarks and our proposed STEM2Code-Eval benchmark. As shown in Tab.\ref{supp:two-stage math} and Tab.\ref{supp:STEM2code}, CodePercept gets consistency improvements.
\begin{table*}[h]
    \centering
    \caption{Performance comparison of various MLLMs across six STEM reasoning benchmarks.}
    \vspace{-1em}
    \label{tab:main_results}
    \scalebox{0.9}{
    \begin{tabular}{l L{14mm}L{13mm}L{13mm}L{13mm}L{13mm}L{15mm} L{13mm}}
        \toprule[1.2pt]
        \multirow{2}{*}{\emph{\textbf{Image Captioner}}} & \multicolumn{6}{c}{Benchmark Datasets (\%)} & {\multirow{2}{*}{Average}} \\
        \cmidrule(lr){2-7}
        & {MathVision} & {MathVista} & {MathVerse} & {DynaMath} & {WeMath} & {LogicVista} & \\
        \midrule[1.2pt]
        \multicolumn{8}{c}{\emph{\textbf{LLM Solver: Qwen3-30A3-Thinking~\cite{yang2025qwen3}}}} \\
        \midrule
        KeyeVL1.5-8B~\cite{team2025kwai} &54.11&64.90&49.95&62.37&33.62&45.19&51.69\\
        Intern-S1-8B~\cite{bai2025intern} &51.67&65.70&51.90&63.61&33.43&51.23 &52.92\\
        GLM-4.1V-9B~\cite{hong2025glm} &53.75&64.60&54.47&66.17&40.76&51.00&55.13\\
        InternVL3.5-8B~\cite{wang2025internvl3} &53.32&67.70&53.40&68.12&41.05&51.68&55.88\\
        MiniCPM-V-4.5~\cite{team2025minicpm4} &53.15&66.60&57.84&65.44&43.71&52.57&56.55\\
        Qwen2.5-VL-72B~\cite{bai2025qwen2} &54.14&67.50&55.40&68.28&44.86&52.34&57.09\\
        Qwen3-VL-30A3B-Instruct~\cite{Qwen3VL_github} &53.59&68.00&66.44&71.67&46.10&53.69&59.92\\
        Claude-Opus 4.1-Thinking~\cite{anthropic2024claude} &59.61  &71.10  &56.19  &73.25  &44.86  &59.28 &60.72 \\
        GPT5-Thinking~\cite{openai2023gpt5} &60.03  &65.20  &69.56  &71.00  &54.57  &53.02  & 62.23 \\
        Qwen3-VL-235A22B-Instruct &60.43&73.80&70.08&77.39&53.05&59.73&65.74\\
        Gemini2.5-Pro &66.80  &74.80  &73.47  &81.42  &60.29  &66.44  &70.53 \\
        \cmidrule(lr){1-8}
        Qwen3-VL-4B-Instruct~\cite{Qwen3VL_github} &54.21 &67.30 &64.59 &69.40 &46.10 &54.14 &59.29 \\
        \rowcolor{cyan!5} \bfseries{CodePercept-4B-S1} &57.63\bbf{+3.4}  &69.60\bbf{+2.3}  &65.59\bbf{+1.0}  &71.38\bbf{+2.0}  &47.81\bbf{+1.7}  &60.40\bbf{+6.3}   & 62.07\bbf{+2.8} \\
        Qwen3-VL-8B-Instruct~\cite{Qwen3VL_github} &54.37  & 69.60 & 63.75 & 72.19 & 45.43 & 56.82 & 60.36 \\
        \rowcolor{cyan!5} \bfseries{CodePercept-8B-S1} &59.31\bbf{+5.0}  &70.20\bbf{+0.6}  &66.52\bbf{+2.8}  &73.20\bbf{+1.0}  &49.14\bbf{+3.7}  &61.52\bbf{+4.7}  &63.32\bbf{+3.0} \\
        Qwen3-VL-32B-Instruct~\cite{Qwen3VL_github} &58.55&72.20&71.09&75.78&48.00&62.19&64.63\\
        \rowcolor{cyan!5} \bfseries{CodePercept-32B-S1} &62.27\bbf{+3.7}  &72.90\bbf{+0.7}  &71.70\bbf{+0.6}  &77.41\bbf{+1.6}  &54.19\bbf{+6.2}  &65.33\bbf{+3.1}  &67.30\bbf{+2.7} \\
        \midrule[1.2pt]
        \midrule[1.2pt]
        \multicolumn{8}{c}{\emph{\textbf{LLM Solver: Qwen3-235A22-Thinking~\cite{yang2025qwen3}}}} \\
        \midrule
        Qwen3-VL-4B-Instruct~\cite{Qwen3VL_github} &59.80 &69.20 &66.39 &71.22 &48.86 &56.82 &62.05 \\
        \rowcolor{cyan!5} \bfseries{CodePercept-4B-S1} &64.71\bbf{+4.9}  &71.30\bbf{+2.1}  &66.73\bbf{+0.3}  &72.40\bbf{+1.2}  &50.00\bbf{+1.1}  &64.65\bbf{+7.8}  &64.97\bbf{+2.9} \\
        Qwen3-VL-8B-Instruct~\cite{Qwen3VL_github} &59.67 &71.00 &63.88 &73.69 &49.14 &58.16 &62.59  \\
        \rowcolor{cyan!5} \bfseries{CodePercept-8B-S1} &66.45\bbf{+6.8}  &71.40\bbf{+0.4}  &67.95\bbf{+4.1}  &75.05\bbf{+1.4}  &52.29\bbf{+3.2}  &62.64\bbf{+4.5}  &65.96\bbf{+3.4} \\
        Qwen3-VL-32B-Instruct~\cite{Qwen3VL_github} &62.66 &74.00 &69.90 &75.54 &56.48 &66.44 &67.50\\
        \rowcolor{cyan!5} \bfseries{CodePercept-32B-S1} &69.96\bbf{+7.3}  &75.90\bbf{+1.9}  &73.56\bbf{+3.6}  &79.50\bbf{+4.0}  &57.81\bbf{+1.3}  &70.02\bbf{+3.6}  &71.13\bbf{+3.6} \\
        \bottomrule[1.2pt]
    \end{tabular}
    }
    \label{supp:two-stage math}
\end{table*}
\begin{table}[h]
    \centering
    \caption{Performance evaluation on our STEM2Code-Eval with 1k samples. We employ three metrics to assess their performances : (1) \textbf{Image Scoring} measures the visual similarity between the generated and original images; (2) \textbf{Code Scoring} assesses the quality, structure, and correctness of the generated Python code itself; (3) \textbf{Exec Rate} reports the execution success rate.
    }
    \vspace{-1em}
    \scalebox{0.9}{
    \begin{tabular}{l@{\hspace{6pt}}c@{\hspace{6pt}}c@{\hspace{6pt}}c@{\hspace{6pt}}c}
    \toprule[1.2pt]
\textbf{Model} & \makecell[c]{\textbf{Image}\\\textbf{Score}} & \makecell[c]{\textbf{Code}\\\textbf{Score}} & \makecell[c]{\textbf{Exec}\\\textbf{Rate}} & \textbf{Avg} \\
        \midrule
        Intern-S1-8B~\cite{bai2025intern} &6.02&15.87&26.60&16.16\\
        InternVL3.5-8B~\cite{wang2025internvl3} &13.50&18.16&56.50&29.38\\
        MiniCPM-V-4.5~\cite{team2025minicpm4} &13.91&23.69&50.80&29.47\\
        MiMo-VL-7B-RL~\cite{xiaomi2025mimo} &14.54&22.41&60.30&32.42\\
        Ovis2.5-9B~\cite{lu2025ovis2} &9.76&11.26&89.40&36.81\\
        KeyeVL1.5-8B~\cite{team2025kwai} &20.33&22.47&73.40&38.73\\
        GLM-4.1V-9B~\cite{hong2025glm} &21.19&26.51&72.00&39.90\\
        Qwen3-VL-4B-Thinking~\cite{Qwen3VL_github} & 25.38 & 34.53 & 75.70 & 45.20 \\
        Qwen2.5-VL-72B-Instruct~\cite{bai2025qwen2} &32.82&25.83&86.30&48.32\\
        Qwen3-VL-8B-Thinking~\cite{Qwen3VL_github} & 29.82 & 41.71 & 78.90 & 50.14 \\
        Qwen3-VL-30A3B-Instruct~\cite{Qwen3VL_github} &33.05&31.04&87.50&50.53\\
        Seed1.6-Vision-nothinking~\cite{guo2025seed1} & 31.22 & 38.56 & 85.50 &51.76 \\
        Qwen3-VL-30A3B-Thinking~\cite{Qwen3VL_github} &37.47&35.53&87.10&53.37\\
        Qwen3-VL-Plus-Instruct~\cite{Qwen3VL_github} & 45.94 & 40.40 & 90.00 & 58.78 \\
        Qwen3-VL-Plus-Thinking~\cite{Qwen3VL_github} & 45.59 & 40.61 & 89.20 & 58.47 \\
        Seed1.6-Vision-Thinking~\cite{guo2025seed1} & 42.03 & 40.74 & 94.70 & 59.15 \\
        Gemini2.5-Flash-Thinking & 57.25 & 60.87 & 85.20 & 67.77 \\
        Claude-Opus 4.1-Thinking & 55.90 & 56.19 & 97.10 & 69.73 \\
        GPT5-Thinking & 64.97 & 64.98 & 96.60 & 75.52 \\
        Gemini2.5-Pro-Thinking & 68.89 & 75.41 & 91.70 & 78.67 \\
        \midrule
        Qwen3-VL-4B-Instruct~\cite{Qwen3VL_github} & 24.55 & 26.42 & 79.40 &43.46 \\
        \rowcolor{cyan!5} \bfseries{CodePercept-4B-S1} & 38.13 & 43.43 & 80.70 & 54.09 \\
        \rowcolor{cyan!10} \bfseries{CodePercept-4B-R1} & \textbf{47.17} & \textbf{45.86} & 91.30 & \textbf{61.44} \\
        \midrule
        Qwen3-VL-8B-Instruct~\cite{Qwen3VL_github} & 28.59 & 28.23 & 85.30 &47.37 \\
        \rowcolor{cyan!5} \bfseries{CodePercept-8B-S1} & 44.53 & 46.78 & 87.60 &\textbf{59.64} \\
        \rowcolor{cyan!10} \bfseries{CodePercept-8B-R1} & \textbf{50.25} & \textbf{47.04} & 93.40 & \textbf{63.56} \\
        \midrule
        Qwen3-VL-32B-Instruct~\cite{Qwen3VL_github} &36.85&39.98&81.80&52.88\\
        \rowcolor{cyan!5} \bfseries{CodePercept-32B-S1} & 61.14 & 56.99 & 93.00 &\textbf{70.38} \\
        \rowcolor{cyan!10} \bfseries{CodePercept-32B-R1} &\bfseries{68.97}  &\bfseries{62.53}  &\bfseries{95.90}  &\textbf{75.80} \\
        \bottomrule
    \end{tabular}
    }
    \label{supp:STEM2code}
\end{table}

\end{document}